\documentclass{article}

\usepackage{microtype}
\usepackage{graphicx}
\usepackage{subcaption}
\usepackage{booktabs} 
\usepackage{tikz}
\usetikzlibrary{fit}
\usetikzlibrary{external}
\usepackage{adjustbox}
\usepackage{enumitem}
\usepackage{hyperref}

\newcommand{\footstar}[1]{$^*$ \renewcommand*\@makefnmark{}\footnotetext{$^\dagger$#1}\makeatother}

\newcommand{\jeo}[1]{\textcolor{purple}{JEO: #1 }}

\newcommand{\zs}[1]{\textcolor{blue}{ZS: #1 }}

\usepackage[preprint]{icml2026}

\usepackage{amsmath}
\usepackage{amssymb}
\usepackage{mathtools}
\usepackage{amsthm}
\usepackage{makecell}
\usepackage{rotating}

\usepackage[capitalize,noabbrev]{cleveref}

\theoremstyle{plain}
\newtheorem{theorem}{Theorem}[section]

\theoremstyle{definition}
\newtheorem{definition}[theorem]{Definition}

\theoremstyle{remark}

\crefname{appendix}{appendix}{appendices}
\usepackage[textsize=tiny]{todonotes}

\icmltitlerunning{Diffeomorphism-Equivariant Neural Networks}

\begin{document}


\twocolumn[
  \icmltitle{Diffeomorphism-Equivariant Neural Networks}



  \icmlsetsymbol{equal}{*}

  \begin{icmlauthorlist}
    \icmlauthor{Josephine Elisabeth Oettinger}{hl,cb} 
    \icmlauthor{Zakhar Shumaylov}{cb} 
    \icmlauthor{Johannes Bostelmann}{hl} 
    \\
    \icmlauthor{Jan Lellmann}{hl} 
    \icmlauthor{Carola-Bibiane Schönlieb}{cb} 
  \end{icmlauthorlist}

  \icmlaffiliation{hl}{Institute of Mathematics and Image Computing, \textbf{University of Luebeck}, Luebeck, Germany}
  \icmlaffiliation{cb}{Department of Applied Mathematics and Theoretical Physics, \textbf{University of Cambridge}, Cambridge, United Kingdom}

  \icmlcorrespondingauthor{Josephine Elisabeth Oettinger}{josephine.oettinger@uni-luebeck.de}

  \icmlkeywords{Machine Learning, ICML} 

  \vskip 0.3in
]



\printAffiliationsAndNotice{}  

\begin{abstract}
Incorporating group symmetries via equivariance into neural networks has emerged as a robust approach for overcoming the efficiency and data demands of modern deep learning.
While most existing approaches, such as group convolutions and averaging-based methods, focus on compact, finite, or low-dimensional groups with linear actions, this work explores how equivariance can be extended to infinite-dimensional groups.  
We propose a strategy designed to induce diffeomorphism equivariance in pre-trained neural networks via energy-based canonicalisation. 
Formulating equivariance as an optimisation problem allows us to access the rich toolbox of already established differentiable image registration methods.
Empirical results on segmentation and classification tasks confirm that our approach achieves approximate equivariance and generalises to unseen transformations without relying on extensive data augmentation or retraining.
\end{abstract}

\section{Introduction}
Deep learning models have become a cornerstone of modern data-driven research in numerous scientific fields~\cite{ML_bio, ML_engin}.
Especially in computer vision~\cite{DL_ComputerVision} and medical imaging~\cite{DL_medIC, DL_medIC2}, deep learning has become a central tool for tasks such as classification~\cite{ML_class}, object detection~\cite{ML_objectDet}, and segmentation~\cite{unet}.
Despite this, their reliance on large-scale, high-quality data and intensive computational requirements remains a major challenge~\cite{DL_book, DLandData}.
Crucially, even with sufficient data, generalisation properties are difficult to guarantee, as they depend heavily on the data seen during training and testing~\cite{DL_datadependency, DL_not_gen}.
\begin{figure*}[ht]
    \begin{center}
    \begin{subfigure}[t]{0.51\textwidth}
        \centering
        \scalebox{0.8}{
        \begin{tikzpicture}
            \node at (0*3.5 -1 ,1.5*3.5) (image-1){\includegraphics[width=2 cm]{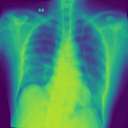}};
            \node[above=-0.1cm of image-1] {\footnotesize \textbf{Input}};
            \node at (1*3.5 +0.3,1.5*3.5) (image0) {\includegraphics[width=2 cm]{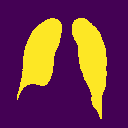}};
            \node[above=-0.1cm of image0] {\footnotesize \textbf{Output}};

            \draw[->, black, thick] (0*3.5 -1 ,1.5*3.5 - 1.1) -- (0*3.5 -1 ,0*3.5+1.25+0.25+1.7) node[pos=0.5, left] { $g$}; 
            \node at (0*3.5-1,0*3.5+1.7) (image3) {\includegraphics[height=2 cm]{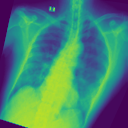}};
            \node[above=-0.1cm of image3] {\footnotesize \textbf{Transformed Input}};
            \node at (1*3.5 +0.3,0*3.5+1.7) (image2) {\includegraphics[height=2cm]{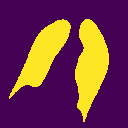}};

            \draw[->, black, thick] (1.5*3.5 -1.4 ,1.5*3.5 - 1.1) -- (1.5*3.5 -1.4 ,0*3.5+1.25+0.25 + 1.7) node[pos=0.5, left] { $g$}; 

            \node[text width=4.5cm ,above=-0.1cm of image2] {\footnotesize \textbf{Output of Transformed Input} }; 

            \draw[->, black, thick] (0*3.5+1.25+0.25 -1.3,1.5*3.5 ) -- (1*3.5-1.25-0.25 +0.5 ,1.5*3.5) node[midway, above] {\footnotesize Segmentation $f_\theta$};
            \draw[->, black, thick] (0*3.5+1.25+0.25 -1.3 ,0*4 + 1.7) -- (1*3.5-1.25-0.25 +0.5 ,0*3.5 + 1.7) node [midway, above] {{\footnotesize Segmentation $f_\theta$}};
        \end{tikzpicture}
        }
        \caption{Diffeomorphism equivariance.}
    \end{subfigure}%
    \hfill
    \vrule width 0.5pt
    \hfill
    \begin{subfigure}[t]{0.47\textwidth}
        \centering
        \scalebox{0.8}{
        \begin{tikzpicture}
            \node at (0*3.5 -1 ,1.5*3.5) (image-1){\includegraphics[width=2 cm]{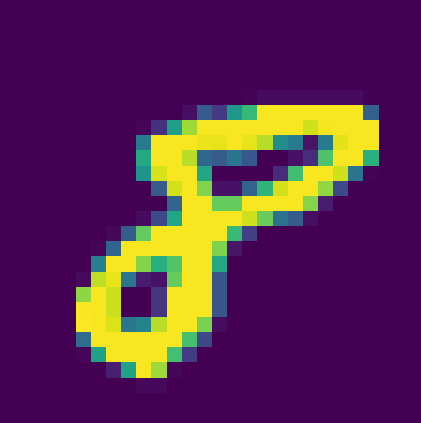}};
            \node[above=-0.1cm of image-1] {\footnotesize \textbf{Input}};

            \draw[->, black, thick] (0*3.5 -1 ,1.5*3.5 - 1.1) -- (0*3.5 -1 ,0*3.5+1.25+0.25+1.7) node[pos=0.5, left] { $g$}; 
            \node at (0*3.5-1,0*3.5+1.7) (image3) {\includegraphics[height=2 cm]{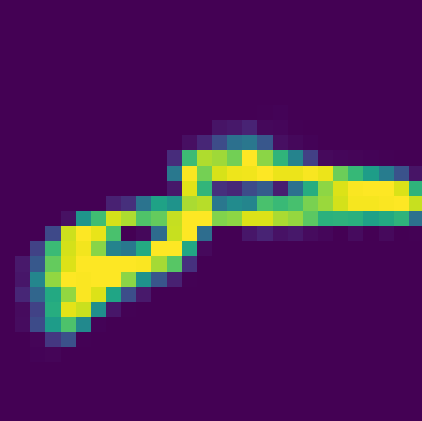}};
            \node[above=-0.1cm of image3] {\footnotesize \textbf{Transformed Input}};
            \node at (1*3.5+0.8,0.75*3.5+1) (image2) { \textbf{``genus two''}};
            \node[draw=black, thick, fit=(image2), inner sep=5pt] (gc_eq_box) {};


            \node[above=0.2cm of image2] {\footnotesize \textbf{Output}}; 

            \draw[->, black, thick] (0*3.5+1.25+0.25 -1.2,1.5*3.5) -- (1*3.5-1.25-0.25 +1 ,0.8*3.5+1) node[midway, above, sloped] {\footnotesize Classification $f_\theta$};
            \draw[->, black, thick] (0*3.5+1.25+0.25 -1.2 ,0*4+1.75) -- (1*3.5-1.25-0.25 +1,0.7*3.5+1) node [midway, above, sloped] {{\footnotesize Classification $f_\theta$}};
        \end{tikzpicture}
        }
        \caption{Diffeomorphism invariance.}
    \end{subfigure}
    \caption{Examples of \textbf{(a)} diffeomorphism equivariance in lung image~\cite{realData} segmentation and \textbf{(b)} diffeomorphism invariance in predictions of topological invariants in MNIST~\cite{mnist}. 
    Applying a diffeomorphism $g \in \mathcal{D}(\mathcal{X})$ to the input should result in a correspondingly transformed segmentation output, i.e., \mbox{$f_\theta(g \cdot x) = g \cdot f_\theta(x)$}, and preserve the classification, i.e.,  \mbox{$f_\theta(g \cdot x) = f_\theta(x)$}. 
    }  
    \label{fig:intro_equivarinace_invariance}
    \end{center}
\end{figure*}
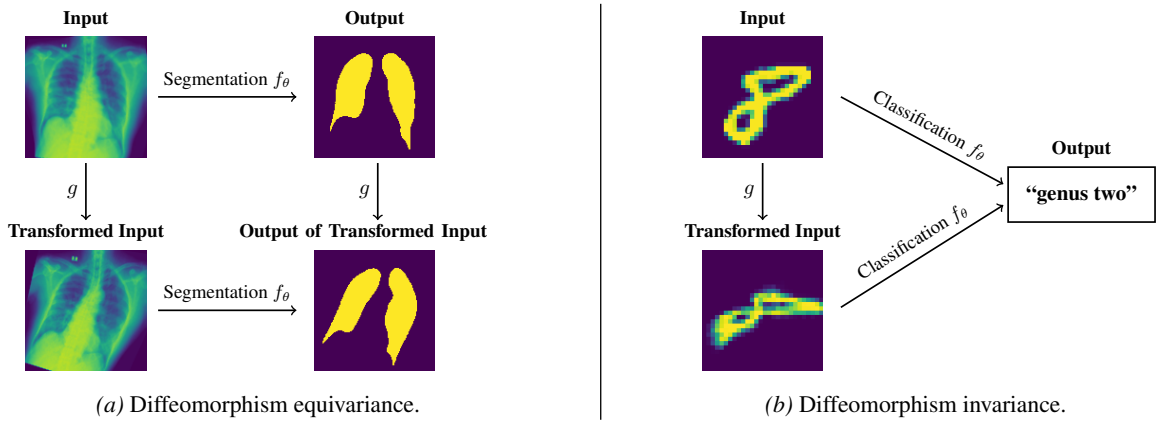

In the quest for more robust and generalisable machine learning models, the focus has shifted towards exploiting known symmetries of the problem through \emph{invariant} or \emph{equivariant neural networks}~\cite{geometric_DL}.
While \emph{group invariance} means that a model’s output remains unchanged when the input is transformed by any element of the group, \emph{group equivariance} means that the model’s output transforms in a predictable way under the group action that is applied to the input.
More formally, let $G$ be a group acting on differentiable manifolds $\mathcal{X}$ and $\mathcal{Y}$. 
A function $f:\mathcal{X} \to \mathcal{Y}$ is called \emph{$G$-invariant}, if for all $g \in G$, and $x \in \mathcal{X}$:
\begin{align}
    f(g \cdot_{\mathcal{X}} x) = f(x),
\end{align} 
and \emph{$G$-equivariant}, if for all $g \in G$, and $x \in \mathcal{X}$:
\begin{align}
    f (g \cdot_{\mathcal{X}} x) = g \cdot_{\mathcal{Y}} f(x), \label{eq:group_equi}
\end{align} 
where the group element $g \in G$ acts on $\mathcal{X}$ and $\mathcal{Y}$ with corresponding actions $\cdot_{\mathcal{X}} $ and $\cdot_{\mathcal{Y}}$. For brevity, we will omit the subscripts when the action is clear from the context.
The distinction between equivariance and invariance is illustrated in \Cref{fig:intro_equivarinace_invariance}, using segmentation and classification as representative tasks under the group of diffeomorphisms. 

Enforcing invariance or equivariance with respect to a relevant transformation group can not only increase robustness but also reduce the need for extensive annotated data and therefore, the training time~\cite{group-eq-CNN, geometric_DL, Example_learning_invariance, scaling_paper}. 
Depending on the task, either group invariance (e.g., classification) or group equivariance (e.g., segmentation) is desired. 
Previous works have primarily focused on small, discrete symmetry groups, in particular rigid motions, or very specific learning settings~\cite{group-eq-CNN, rot-inv-CNN}.
However, in many applications, more generic non-linear transformations are desirable.
These transformations are naturally modelled as \emph{diffeomorphisms}.
\begin{definition}[Diffeomorphism]
    Let $\mathcal{X}$ and $\mathcal{Y}$ be differentiable manifolds.
    A map $g:\mathcal{X}\to\mathcal{Y}$ is called a \emph{diffeomorphism}, if it is bijective, differentiable, and has a differentiable inverse $g^{-1}: \mathcal{Y} \to \mathcal{X}$.
    The set of all such maps is denoted as $\mathcal{D}(\mathcal{X}, \mathcal{Y})$.
\end{definition}
\noindent If $\mathcal{X} = \mathcal{Y}$, this set forms an infinite-dimensional group under composition, which we call $\mathcal{D}(\mathcal{X})$~\cite{DiffeoGroup}.
Diffeomorphisms are frequently used in imaging applications, particularly medical image registration~\cite{originalLDDMM, VoxelMorph, DARTEL}, where they model realistic anatomical transformations while preserving topological properties~\cite{diffeos-medical, Med_IR}.
A neural network that is equivariant to diffeomorphisms would therefore be highly valuable for real-world tasks, e.g., in cell segmentation in histopathology with variability in cell shapes or lung segmentation with different breathing phases~\cite{diffeo_cell, diffeo_lung}.

To formalise our setting, let $\mathcal{X}$ denote the space of greyscale images defined on the domain $\Omega \subseteq \mathbb{R}^2$ with intensities in $[0,1]$, modelled as functions that map from $\Omega$ to $[0, 1]$, 
\begin{equation}
    \mathcal{X} := \{x \;|\; x:\Omega \to [0, 1]\}.
\end{equation}
The group of diffeomorphisms $\mathcal{D}(\mathcal{X})$ acts on the image domain $\Omega$. 
We define the group action of $\mathcal{D}$ on $\mathcal{X}$ as
\begin{equation}
    (g \cdot x) (p) := (x \circ g)(p) = x (g(p))
\end{equation}
for all $p \in \Omega$, $x \in \mathcal{X}$ and $g \in \mathcal{D}(\mathcal{X})$.
This group action produces deformed images as, e.g., the input in Figure~\ref{fig:intro_set_up}.

\subsection{Contribution}

In this work, we introduce Diffeomorphism-Equivariant Neural Networks (DiffeoNN)\footnote{Source code will be made available upon publication.}, a framework for adapting pre-trained models to achieve diffeomorphism equivariance. Our approach extends the energy canonicalisation method of \citet{ZaksLieLAC, can_main} from finite-dimensional Lie groups to the infinite-dimensional group of orientation-preserving diffeomorphisms. Unlike standard equivariance methods, requiring integration over locally compact groups, DiffeoNN utilises group-agnostic canonicalisation to achieve equivariance without exhaustive sampling. We parametrise these transformations using \emph{stationary velocity fields (SVFs)}, enabling the framework to handle a broad class of complex spatial deformations, while being computationally feasible.

\begin{figure}[ht]
    \centering
    \hspace{1cm}
    \scalebox{0.6}{
    \begin{tikzpicture}
\node at (0*3.5 -1 ,2*3.5) (image-1){\includegraphics[width=2.5 cm]{graphics/lung_bench/26_moving.png}};
\node[above=-0.1cm of image-1] { Input};
\node at (2*3.5 +1,2*3.5) (image0) {\includegraphics[width=2.5 cm]{graphics/lung_bench/26_gt_mask.png}};

\node[above=-0.1cm of image0] { Ground Truth};

\node at (2*3.5 +1,3.5*1) (image1) {\includegraphics[width=2.5 cm]{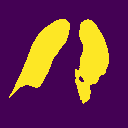}};
\node[ above=-0.1cm of image1] {Na\"ive}; 
\node at (0*3.5-1,0*3.5) (image3) {\includegraphics[width=2.5 cm]{graphics/lung_bench/26_warped.png}};
\node[text width=3cm, above=-0.1cm of image3] {Canonicalised Input};
\node at (3.5*1,0*3.5) {\includegraphics[width=2.5cm]{graphics/lung_bench/26_warped_mask.png}};
\node at (2*3.5+1,0*3.5) (image2) {\includegraphics[width=2.5 cm]{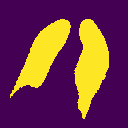}};

\node[above=-0.1cm of image2] {Ours}; 

\draw[->, black, thick] (0*3.5+1.25+0.25 -1,2*3.5) -- (2*3.5-1.25-0.25 +1 ,2*3.5) node[midway, above] { True Segmentation};
\draw[->, black, thick] (0*3.5 -1 ,2*3.5 - 1.25 - 0.25) to[out=310,in=180] (2*3.5-1.25-0.25 +1,3.5*1) ;
\node at (3.5*1 ,3.5*1 + 0.45) {Network $f_\theta$}; 

\draw[->, black, thick] (0*3.5 -1 ,2*3.5 - 1.25 - 0.25) -- (0*3.5 -1 ,0*3.5+1.25+0.25 + 0.35) node[pos=0.5, left] (gcdef) { }; 

\draw[->, black, thick] (0*3.5+1.25+0.25 -1 ,0*4) -- (3.5*1-1.25-0.25,0*3.5) node [text width=0.5cm, midway, above] {{$f_\theta$}}; 
\draw[->, black, thick] (3.5*1+1.25+0.25,0*3.5) -- (2*3.5-1.25-0.25 +1 ,0*3.5) node[midway, above] (gcdef2){$g_x^{-1}$};

\node at (0*3.5 +1.2, 1*3.5 - 0.9) (gc_eq) {\color{black}\scalebox{0.95}{ $g_x \in \underset{g \in \mathcal{D}_{\text{SVF}}(\mathcal{X})}{\operatorname{arg\;min}}\; E_{\text{can}}\Big(\adjustbox{valign=c}{\includegraphics[width=0.5cm]{graphics/lung_bench/26_moving.png}}, g\Big)$}}; 

\node[draw=black, thick, fit=(gc_eq), inner ysep=0.pt,inner xsep=-2pt] (gc_eq_box) {};
\end{tikzpicture}
}
    \caption{Canonicalisation (DiffeoNN) for lung segmentation.
    The network $f_\theta$ is pre-trained on a simple chest X-ray/ground-truth segmentation dataset.
    Applying the trained model $f_\theta$ to a diffeomorphically transformed image without canonicalisation results in an Intersection-over-Union (IoU) of $0.8769$ \textbf{(na\"ive)} with clearly visible errors in the segmentation map.
    Using canonicalisation, i.e., transforming the input closer to the training dataset before applying the network $f_\theta$ and reversing the transformation, achieves an almost perfect segmentation with an IoU of $0.9560 $ \textbf{(ours)}.
    }
    \label{fig:intro_set_up}
\end{figure}

Extending the energy canonicalisation framework (\Cref{sec:framework}) to an infinite-dimensional setting necessitates two primary technical shifts inspired by classical image registration \cite{originalLDDMM, Younes}. First, we propose a new energy functional (\Cref{sec:can_energy}) that integrates a Variational Autoencoder (VAE) loss with an adversarial discriminator \cite{advReg1, advReg2} and a deformation regulariser \cite{han2023diffeomorphic, johannes} to ensure structural consistency in the canonicalised outputs. Second, we replace the Lie-algebra optimisation used in LieLAC \citep{ZaksLieLAC} with a gradient-based strategy specifically tailored for SVFs \cite{johannes}. These adaptations provide a generalised framework applicable across various computer vision tasks, such as the segmentation example shown in \Cref{fig:intro_set_up}.

To evaluate our DiffeoNN, we perform several experiments for the tasks of \emph{diffeomorphism-equivariant segmentation} (\Cref{sec:exp_segmentation}) utilising a {synthetic dataset} of nested squares and a {real-world dataset of chest X-rays}~\cite{realData}, as well as \emph{diffeomorphism-invariant homology classification} (\Cref{sec:exp_classification}), using the {MNIST dataset}~\cite{mnist}. Our results demonstrate that DiffeoNN consistently outperforms the naïve baseline, being on par with the data augmentation approach, while resulting in consistently fewer outliers. We further verify that the canonicalisation step is diffeomorphism-invariant by construction, supporting this claim through visual comparisons and energy-level analysis of transformed image pairs.


\subsection{Related Work}


Unlike finite groups or simple transformations such as rotations and translations, the group of diffeomorphisms is infinite-dimensional. Consequently, it cannot be compactly parametrised for integration into standard equivariance approaches, like data augmentation~\cite{DataAugmentation}, group-equivariant layers~\cite{group-eq-CNN}, and averaging-based methods~\cite{group_averaging, frame_averaging}. Given the limitations of these approaches, \emph{canonicalisation} emerges as the most viable strategy for achieving diffeomorphism equivariance.


\paragraph{Data augmentation}
Data augmentation~\cite{DataAugmentation} offers a heuristic framework to approximate diffeomorphism equivariance~\cite{diffeoAug} by expanding the training set through random group transformations. While conceptually viable, this approach encounters significant computational bottlenecks, primarily due to the necessity of sampling and transforming data within an infinite-dimensional space. As finite sampling cannot capture the full variability of the diffeomorphism group, this approach fails to provide rigorous guarantees of formal equivariance. Despite these limitations, it represents the primary alternative to canonicalisation within existing literature; we therefore adopt it as our comparative baseline.

\paragraph{Group-Equivariant Layers}
In contrast to data-based methods, incorporating symmetries through equivariant architectures can guarantee group equivariance by design.
Convolutional neural networks (CNNs) are famously translation-equivariant due to the weight sharing in their convolutional layers~\cite{inv_CNNs1, inv_CNNs2}.
It has been suggested that this is an important reason for the efficiency of CNNs in tasks such as image classification~\mbox{\cite{why_Cnns,why_CNNS2}}. 
This has motivated architectures with group equivariance with respect to other symmetries incorporated directly in the network layers, such as rotation invariance~\cite{rot-inv-CNN}. 

This idea was generalised to arbitrary finite groups via Group Convolutional Networks (G-CNNs)~\cite{group-eq-CNN}, where the standard convolutions are replaced with \emph{group convolutions} over discrete symmetry groups.
This setup generally assumes ``nice'' groups acting in representations, i.e. linearly on the underlying vector space.
G-CNNs have been extended to compact groups such as $SO(2)$ and $SO(3)$ via spherical CNNs~\cite{spherical_G-cnn}, and to efficient implementations using fast Fourier transforms on groups~\cite{comp_G-cnn}.

While being a promising strategy for incorporating equivariance into neural networks, group convolutions require the group $G$ to be finite (or at least compact and tractably parametrised~\cite{group-eq-CNN}), being implemented by a discrete sum over $G$. 
This makes G-CNNs incompatible with continuous and infinite groups, or those not acting in representations, like the group of diffeomorphisms.

Building on G-CNNs, \emph{Lie group convolutions} generalise to arbitrary Lie groups via integration with respect to the Haar measure. 
Various approaches have been proposed to make them computationally tractable, such as using different parametrisations~\cite{Lie-G-cnn, spline-g-cnn}, exploiting the local Lie group structure~\cite{lie-alg-g-cnn}, and group-specific approaches for roto-translation groups~\cite{roto-g-cnn}.
While Lie group convolutions cover continuous and infinite groups, they do not apply to our setting involving the group of diffeomorphisms, as they require the existence of a Haar measure~\cite{diffeoNoHaar}.


\paragraph{Averaging-based Approaches}
Averaging methods offer an alternative way to achieve equivariance without redesigning the internal architecture of a neural network. Instead of building specific symmetries into the layers, these methods operate on the \emph{group orbit} of input $x$, defined as the set of all elements that can be reached from $x$ by the action of all elements of a group $G$:
\begin{equation}
    \mathcal{O}_x := \{ g \cdot x \mid g \in G \}.
\end{equation}
These strategies range from group averaging~\cite{group_averaging}, which integrates over the entire orbit, to frame averaging~\cite{frame_averaging} and weighted frames~\cite{can_discont}, which utilise input-dependent subsets or measures to improve tractability. Despite these refinements, the requirement to integrate over or sample from the group remains a bottleneck for diffeomorphisms, where the absence of a Haar measure makes the definition of a suitable integration measure computationally infeasible.

\paragraph{Canonicalisation}

An approach closely related to, but distinct from frame averaging, is \emph{orbit canonicalisation}~\cite{can_main}. Canonicalisation maps each input $x \in X$ to a \emph{canonical representative} $x_c$ within its orbit $\mathcal{O}_x$. By restricting the model's domain to these representatives, the task complexity is significantly reduced, allowing standard architectures to handle large or continuous groups. This approach is highly efficient~\cite{can_gen_bounds}, robust to approximate symmetries~\cite{can_main}, with expressivity easily generalisable through probabilistic symmetry breaking~\cite{lawrence2025detecting,Symmetry_Breaking}.

A notable extension of this paradigm is Lie Algebra Canonicalization (LieLAC)~\cite{ZaksLieLAC}, framing the \emph{canonicalisation step} as a pre-processing step. Crucially, it can be realised without explicitly equivariant architectures, by defining the canonical representative as the minimiser of a task and data dependent energy function $E:X \to [0,\infty]$ over the group orbit: 
\begin{equation}
    c(x) := \arg\min_{y \in \mathcal{O}_x} E(y).\label{eq:canon-idea}
\end{equation}
Intuitively, the energy $E$ is chosen such that its minimisers are ``close'' to the training data~$X_E$. LieLAC utilises Lie algebra descent to find this minimiser, which combined with a pretrained model trained on sparse data $X_E$, enables equivariance for arbitrary Lie groups without modifications of the underlying model. However, LieLAC is inherently restricted to finite-dimensional Lie groups, leaving the infinite-dimensional case as an open problem. In this work, we directly solve this by extending energy canonicalisation to the problem of diffeomorphism equivariance.

\section{Framework}\label{sec:framework}


Given a data manifold $\mathcal{X}$,
the central idea of DiffeoNN is to train a simple model~$f_\theta$ only on a sparse labelled training dataset $X_E \subseteq \mathcal{X}$ and then -- without additional training -- to extend it to perform well on the entire dataset $X \subseteq \mathcal{X}$ containing all diffeomorphic transformations of $X_E$.
This also allows us to make pretrained networks equivariant without retraining.
In analogy with~\cite{VoxelMorph}, we parametrise the diffeomorphisms using \emph{stationary velocity fields}~(SVFs).
We call this set of \emph{SVF-generated diffeomorphisms} \mbox{$\mathcal{D}_{\text{SVF}}(\mathcal{X}) \subset \mathcal{D}(\mathcal{X})$.}

Inspired by \citet{ZaksLieLAC}, our extended model $\Tilde{f}_\theta$ consists of three steps: a \emph{canonicalisation step}, an \emph{inner model trained on the training data $X_E$}, and a \emph{reverse canonicalisation step} (Figure~\ref{fig:mymethod_setup}):
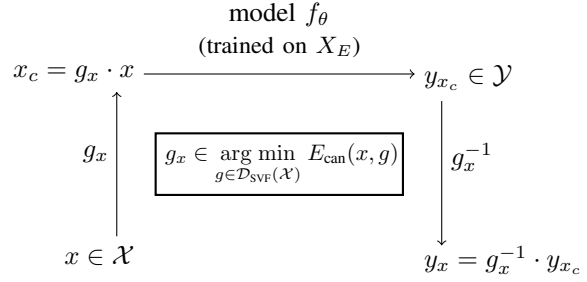
\begin{figure}[t]
\centering
\begin{tikzpicture}[scale=1.2]
\useasboundingbox (-1.6,0) rectangle (5,3.5);

  \coordinate (A) at (0,2.7);
  \coordinate (B) at (3,2.7);
  \coordinate (C) at (0,0.7);
  \coordinate (D) at (3,0.7);

  \draw[->] (0,2.5) -- node [text width=4cm, midway, above=0.1cm, align=center] {model $f_\theta$ \\ {\footnotesize (trained on $X_E$)}} (3,2.5);
  \draw[->] (-0.3,0.7) -- node [text width=0.9cm, midway, above=-0.1cm] {$g_x$} (-0.3,2.3);
  \draw[->] (3.3,2.2) -- node [text width=1.4cm, midway, above=-0.1cm, above right] {$g_x^{-1}$} (3.3,0.6);
  \node at (1.5, 1.5) (gc_eq) {\color{black}\scalebox{0.85}{ $g_x \in \underset{g \in \mathcal{D}_{\text{SVF}}(\mathcal{X})}{\operatorname{arg\;min}}\; E_{\text{can}}(x,g)$}}; 

\node[draw=black, thick, fit=(gc_eq), inner ysep=0.pt,inner xsep=-2pt] (gc_eq_box) {};

  \node[below left] at (A) {$x_c = g_x \cdot x$};
  \node[below right] at (B) {$y_{x_c} \in \mathcal{Y}$};
  \node[below left] at (C) {$x \in \mathcal{X}$};
  \node[below right] at (D) {$y_x = g_x^{-1} \cdot y_{x_c}$};
\end{tikzpicture} 
    \caption{
    In the \textbf{canonicalisation step}, an input $x \in \mathcal{X}$ is canonicalised to $x_c = g_x \cdot x$, where $g_x$ is a canonicalising element that is determined by solving a task-specific optimisation problem. 
    Then the \textbf{inner model $f_\theta$} is applied to $x_c$.
    Its output $y_{x_c} := f_\theta(x_c)$ is transformed by the \textbf{reverse canonicalisation} $g_x^{-1}$ to obtain the final output $y_x = g_x^{-1} \cdot y_{x_c}$.
    } 
    \label{fig:mymethod_setup}
\end{figure}

\begin{enumerate}[leftmargin=*]
    \item In the \textbf{canonicalisation step}, we find a representation $x_c$ for a given input $x \in X$ such that $x_c$ is ``close'' to the training data $X_E$.
    A \emph{canonicalising element} (diffeomorphism) $g_x \in \mathcal{D}_{\text{SVF}}(\mathcal{X})$ is found by minimising a task-specific canonicalisation energy: 
    \begin{equation}
        g_x \in \underset{g \in \mathcal{D}_{\text{SVF}}(\mathcal{X})}{\operatorname{arg\;min}}\; E_{\text{can}}(x,g).\label{eq:canon-svf}
    \end{equation} 
    The canonicalised input is then $x_c := g_x \cdot x$.
    \item At the core is a \textbf{model $f_\theta: \mathcal{X} \to \mathcal{Y}$}, $x\mapsto f_\theta(x)=:y_x \in \mathcal{Y}$\, that is \textbf{trained on the smaller dataset~$X_E$}. The range $\mathcal{Y}$ is a task-specific output space such as a set of labels or segmentation masks.
    The model performs the desired task 
    and outputs $y_{x_c} = f_\theta(x_c) \in \mathcal{Y}$\, for the canonicalised input $x_c$.
    \item To obtain the final output, a \textbf{reverse canonicalisation} is applied: 
    \begin{equation}
        \Tilde{f}_\theta(x) := g_x^{-1} \cdot (f_\theta (g_x \cdot x)),\label{eq:reverse-canon}
    \end{equation}
    where $g_x^{-1} \in \mathcal{D}_{\text{SVF}}(\mathcal{Y})$ reverses the map corresponding to the canonicalising element~$g_x$ on $\mathcal{Y}$. 
    For example, if $\mathcal{X}$ and $\mathcal{Y}$ are both image spaces under the usual diffeomorphism action, $g_x^{-1}$ would apply the inverse of $g_x$ in order to obtain equivariance, while the trivial action in the $\mathcal{Y}$ space would result in invariance.
\end{enumerate}
Such a process has been shown to result in an equivariant model \citep{can_main,ZaksLieLAC}, i.e., $\Tilde{f}_\theta(h\cdot x) = h\cdot \Tilde{f}_\theta(x)$ for all $h\in G$, whenever for all $h\in G$ the energy satisfies $E_{\text{can}}(h\cdot x, g) = E_{\text{can}}(x,g\cdot h)$. This turns out to be powerful, as for arbitrary $E:\mathcal{X}\to\mathbb{R}$, the function $E(g\cdot x)$ satisfies the rule above. 


\section{Canonicalisation for Diffeomorphisms}\label{sec:canonincalisation}
Applying the canonicalisation framework to the group of diffeomorphisms $\mathcal{D}(\mathcal{X})$ requires defining the canonicalisation step \eqref{eq:canon-idea}. In practice, this entails parametrising $\mathcal{D}(\mathcal{X})$ and finding a suitable energy $E_{\text{can}}$ that is small whenever $g \cdot x$ is ``close'' to the training data $X_E$, with $g$ physically plausible, e.g., orientation-preserving, diffeomorphisms.


\subsection{Parametrisation via SVFs}\label{sec:DiffeoNN_SVF_diff}
As an approximate parametrisation of the group of diffeomorphisms $G= \mathcal{D}(\mathcal{X})$, we consider the structured and computationally efficient subset of stationary velocity field (SVF)-generated diffeomorphisms $\mathcal{D}_{\text{SVF}}(\mathcal{X})$.
Such transformations are commonly used in deformable image registration due to their smoothness, easily available inverses, and well-defined theoretical structure \cite{originalLDDMM, DARTEL, VoxelMorph, Younes}.

\begin{definition}[Stationary Velocity Field (SVF)]
    A \emph{ stationary velocity field} is defined as a vector field \mbox{$v:\Omega  \to \mathbb{R}^d$,} where $v(p)$ specifies the velocity 
    at position $p$. 
\end{definition}
\noindent A velocity field $v$ can be used to generate a \emph{flow} $\varphi$:
\begin{definition}[Flow]
    Given a stationary velocity field $v$, the associated \emph{flow} $\varphi: \Omega \times [0,1] \to \Omega, \varphi_t(x) := \varphi(x,t)$ is defined as the solution to the ordinary differential equation (ODE) with $t\in [0,1]$ and $p \in \Omega$:
    \begin{equation}
        \frac{\mathrm{d} \varphi_t(p)}{\mathrm{d}t}= v(\varphi_t(p)), \quad \varphi_0(p)=p. \label{eq:ode_diffeo}
    \end{equation}
\end{definition}
\noindent 
For $v \in \mathcal{C}_0^1(\Omega, \mathbb{R}^d)$, the solution of the ODE \eqref{eq:ode_diffeo} at $t=1$ yields a diffeomorphism \mbox{$ g:= \varphi_1 $}~\cite{svf_creation}. %

The solution map is defined by $\exp(v):=\varphi_1 =g$, called the \emph{exponential map},
and connects the set of smooth vector fields to the group of diffeomorphisms \cite{Younes, DiffIA} similarly to the exponential map that connects a Lie algebra to its Lie group.
Compared to the time-dependent more general LDDMM formulation~\cite{originalLDDMM}, SVF-generated diffeomorphisms are computationally more efficient, as the temporal integration can be performed by scaling and squaring~\cite{ScalingAndSquaring}.
While not every orientation-preserving diffeomorphism can be generated by a SVF~\cite{hernandez08}, $\mathcal{D}_{\text{SVF}}$ suits our method as it includes $\mathrm{id}:=\exp(0)$, and is closed under inverses, as $g^{-1} = \exp(v)^{-1} = \exp(-v)$.

\begin{figure*}[ht]
    \centering
    \begin{tabular}{ccccccccc}
        \textbf{\footnotesize \makecell{Input $x$}} & &  \textbf{\footnotesize \makecell{Canonicalised \\ Input $x_c$}} & \textbf{\footnotesize \makecell{Segmentation \\ of $x_c$}} & & \textbf{\footnotesize \makecell{Output \\ DiffeoNN \\ (Ours)}} & \textbf{\footnotesize \makecell{Output \\ Na\"ive U-Net}} & \textbf{\footnotesize \makecell{Output \\ Augmented \\ U-Net}} & \textbf{\footnotesize \makecell{Ground-Truth \\ Segmentation}} \\

        \includegraphics[width=0.09\textwidth]{graphics/lung_bench/26_moving.png} & &
        \includegraphics[width=0.09\textwidth]{graphics/lung_bench/26_warped.png} &
        \includegraphics[width=0.09\textwidth]{graphics/lung_bench/26_warped_mask.png} & &
        \includegraphics[width=0.09\textwidth]{graphics/lung_bench/26_pred_mask.png} &
        \includegraphics[width=0.09\textwidth]{graphics/lung_bench/26_compare_mask.png} &
        \includegraphics[width=0.09\textwidth]{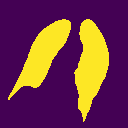} &
        \includegraphics[width=0.09\textwidth]{graphics/lung_bench/26_gt_mask.png} \\
    
        \includegraphics[width=0.09\textwidth]{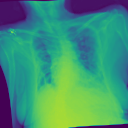} & &
        \includegraphics[width=0.09\textwidth]{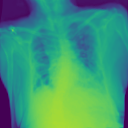} &
        \includegraphics[width=0.09\textwidth]{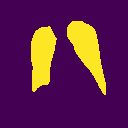} & &
        \includegraphics[width=0.09\textwidth]{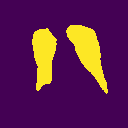} &
        \includegraphics[width=0.09\textwidth]{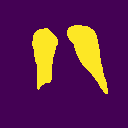} &
        \includegraphics[width=0.09\textwidth]{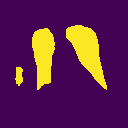} &
        \includegraphics[width=0.09\textwidth]{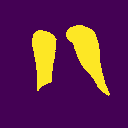} \\

        \includegraphics[width=0.09\textwidth]{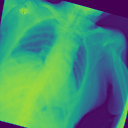} & &
        \includegraphics[width=0.09\textwidth]{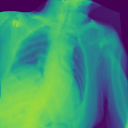} &
        \includegraphics[width=0.09\textwidth]{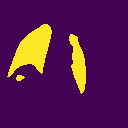} & &
        \includegraphics[width=0.09\textwidth]{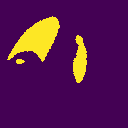} &
        \includegraphics[width=0.09\textwidth]{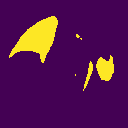} &
        \includegraphics[width=0.09\textwidth]{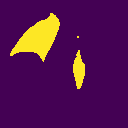} &
        \includegraphics[width=0.09\textwidth]{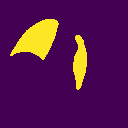} \\

    \end{tabular}
    \caption{Lung segmentation of diffeomorphically transformed chest X-ray images from~\cite{realData}.
    Shown are the steps of DiffeoNN \textbf{(column two and three)}, as well as the final outputs of DiffeoNN, an augmented U-Net, and the inner U-Net of DiffeoNN without augmentation (na\"ive) \textbf{(column four, five, and six).}
    DiffeoNN produces more accurate lung segmentations than the na\"ive approach. While segmentations of DiffeoNN and augmented U-Net are close to the ground truth, the augmented do show some small artefacts. 
    }
    \label{fig:exp_lung}
\end{figure*}
\subsection{Canonicalisation Energy} \label{sec:can_energy}
The canonicalisation energy $E_{\text{can}}$ measures the similarity of a transformed image with the training data and encourages more ``realistic'' transformations within the allowed set of transformations $\mathcal{D}_{\text{SVF}}(\mathcal{X})$. 
In this work, the energy $E_{\text{can}}: X \times \mathcal{D}_{\text{SVF}}(\mathcal{X}) \to \mathbb{R}$ consists of an image similarity energy $E_{X_E}: X \to \mathbb{R}$ and a regulariser $E_{\text{reg}}: \mathcal{D}_{\text{SVF}}(\mathcal{X}) \to \mathbb{R}$:
\begin{align}
    &E_{\text{can}}(x, g) : =   E_{X_E} (g \cdot x)  + E_{\text{reg}}(g). \nonumber
\end{align}
It is important to note that inclusion of a regularisation term technically violates the conditions required for canonicalisation to remain strictly equivariant \citep{can_main, ZaksLieLAC}. Nevertheless, this approach is justified by two primary considerations. Firstly, from a practical standpoint, the space $\mathcal{D}_{\text{SVF}}(\mathcal{X})$ is still much too large; the transformations of interest are often ``small.'' Thus, $\|\nabla v\|_2^2$ in \eqref{eq:E_reg} functions as a soft domain constraint, ensuring the optimisation remains well-posed without strictly restricting the transformation space. Secondly, the fundamental objective of canonicalisation is equivariance of $\Tilde{f}_\theta (x) = g_x^{-1} \cdot {f}_\theta (g_x \cdot x)$ itself, for which we only require that the task admits such equivariance, and data is in-domain: ${f}_\theta (g_x \cdot x) \approx g_x \cdot {f}_\theta (x)$. Thus, approximate equivariance is guaranteed provided that the minimisers are attainable.
\paragraph{Image Similarity Energy}
The image similarity energy estimates the similarity of an input to the training dataset $X_E$, defined for positive scalars $\lambda_{\text{VAE}}$ and $\lambda_{\text{adv}}$ as:
\begin{equation}
    E_{X_E}(g \cdot x) := \lambda_{\text{VAE}} E_{\text{vae}} (g \cdot x) + \lambda_{\text{adv}} E_{\text{adv}} (g \cdot x), \label{eq:E_XE}
\end{equation}
The \textbf{VAE-based energy} $E_\text{VAE}$ is the \emph{variational autoencoder (VAE)} loss~\cite{vae}, trained on the training dataset $X_E$.
Low energy means the input can be well reconstructed by the VAE. 
The \textbf{adversarial energy} $E_{\text{adv}}$ is based on an adversarial discriminator 
trained to distinguish between the training images $X_E$ (``real'') and diffeomorphically transformed ones (``fake'') following~\cite{advReg1, advReg2, ZaksLieLAC}.
The ``fake'' diffeomorphically transformed images are generated by randomly sampling deformation parameters from a radial basis function parametrisation \cite{Forti16032014} to warp the training images $X_E$. $E_\text{adv}$ is then the adversarial energy of the trained discriminator, and is smaller for $x$ that are ``closer'' to the training dataset $X_E$. 
Crucially, training of this discriminator is both stable and computationally efficient, requiring minimal overhead.

\paragraph{Regularising Energy}
In order to encourage physically plausible transformations,
we follow~\cite{han2023diffeomorphic, johannes} and define the \textbf{regulariser}~$E_{\text{reg}}$ as
\begin{align}
    E_{\text{reg}}(g) := &\underbrace{ \lambda_{\Delta} \|\nabla v\|_{2}^2}_{\text{gradient loss}} 
    + \underbrace{ \lambda_{\mathcal{J}} \| \max(0, - \det (\mathcal{J}_g))\|_1}_{\text{Jacobian determinant loss}}, \label{eq:E_reg} 
\end{align}
where $\lambda_{\Delta}$ and $\lambda_{\mathcal{J}}$ are positive scalar weights, $v$ the SVF that induces $g$, and $\det (\mathcal{J}_g)$ the Jacobian determinant of $g$. 
The gradient loss is large for deformations with high spatial variation in their flow field. 
Therefore, it is small when the deformation is smooth.
The Jacobian determinant loss penalises negative Jacobian determinants. 
The intent is to enforce $\det (\mathcal{J}_g (p)) > 0 $ for all $p \in \Omega$, as such diffeomorphisms are orientation-preserving, ensuring physical plausibility. 
A negative determinant indicates local reversal of orientation, such as folding or tearing, undesirable in most image processing tasks.
Therefore, restricting to diffeomorphisms with a positive Jacobian determinant is a common practice in image processing and has been widely researched~\cite{jacDet1,jacdet2}.


\subsection{Canonicalisation via Gradient-based Optimisation} \label{sec:Johannes}
It remains to solve the canonicalisation step \eqref{eq:canon-svf}.
In contrast to the approach of LieLAC~\cite{ZaksLieLAC}, which uses Lie algebra descent, we 
use a gradient-based optimisation method similar to \cite{VoxelMorph, VoxelMorph2} to minimise the canonicalisation energy $E_{\text{can}}$. We follow~\cite{han2023diffeomorphic, johannes} in using 
a \emph{Sinusoidal Representation Network (SIREN)} ~\cite{SIREN} to parametrise the SVFs and a \emph{Scaling and Squaring} approach~\cite{ScalingAndSquaring} to compute the solution $g_\theta = \exp (v_\theta)$ of the ODE.


Note that the canonicalisation step can only ever be approximately diffeomorphism-invariant in practice. 
Exact minimisers are numerically inaccessible due to the finite iteration budget of the gradient-based optimisation scheme~\cite{johannes}, while inclusion of regularisation breaks strict equivariance. Nevertheless, these effects remain sufficiently small in practice to be negligible, allowing the overall network $\Tilde{f}_\theta$ to maintain a high degree of approximate equivariance by construction.

\begin{figure*}[ht]
    \centering
    \begin{tikzpicture}
    \node (table) {
    \begin{tabular}{cccccccc}
        \textbf{\tiny $x \in X_{E}$} \hspace{-6pt} & \textbf{\tiny $x_c$} \hspace{-16pt} &  \textbf{\tiny $x \in X_{E}$} \hspace{-16pt} & \textbf{\tiny $x_c$} \hspace{-16pt} & \textbf{\tiny $x \in X_{E}$} \hspace{-16pt} & \textbf{\tiny $x_c$} \hspace{-16pt} & \textbf{\tiny $x \in X_{E}$} \hspace{-16pt} & \textbf{\tiny $x_c$} \hspace{-16pt}  \\
        \hspace{-16pt}
        \includegraphics[width=0.13\textwidth]{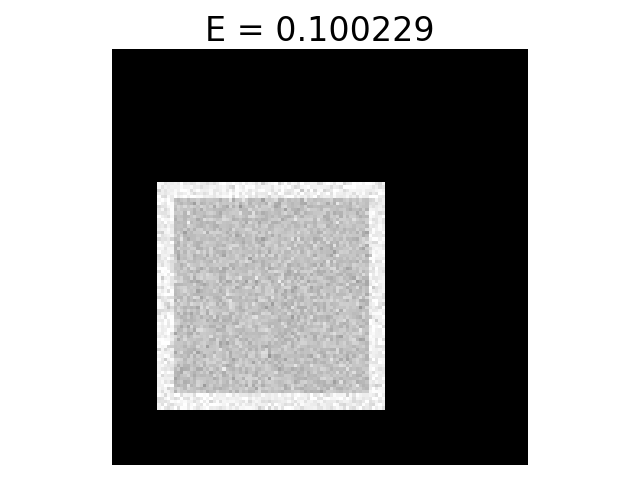} \hspace{-16pt} &
        \includegraphics[width=0.13\textwidth]{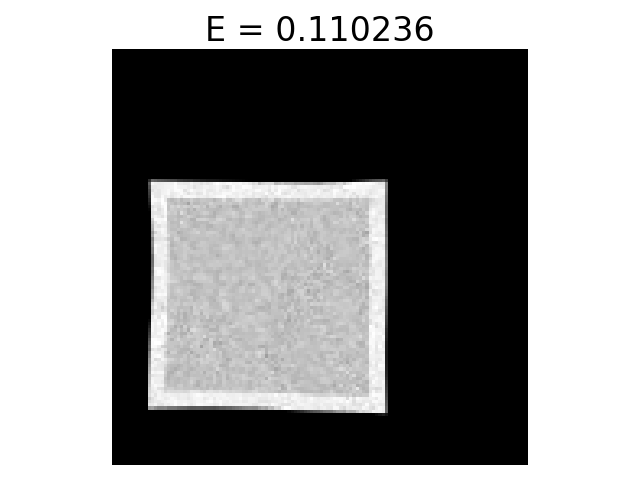} \hspace{-16pt} &
       \includegraphics[width=0.13\textwidth]{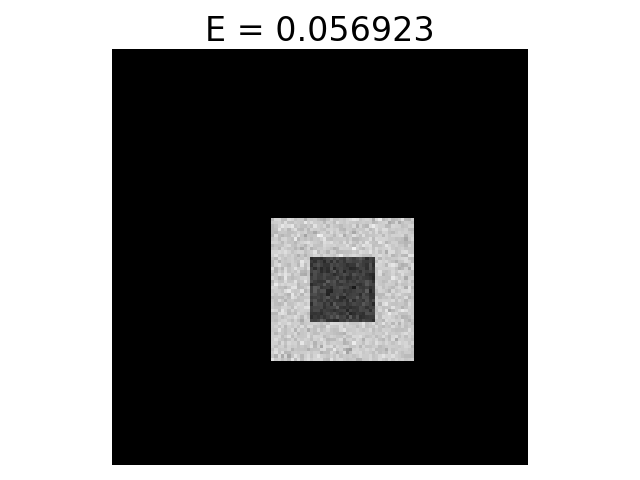} \hspace{-16pt} &
        \includegraphics[width=0.13\textwidth]{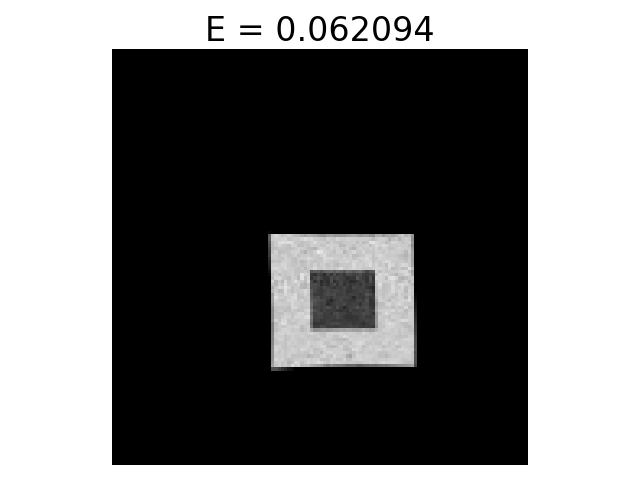} \hspace{-16pt} &
        \includegraphics[width=0.13\textwidth]{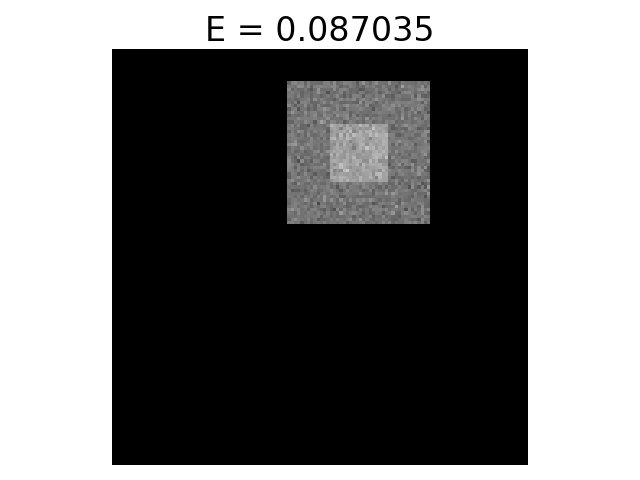} \hspace{-16pt} &
        \includegraphics[width=0.13\textwidth]{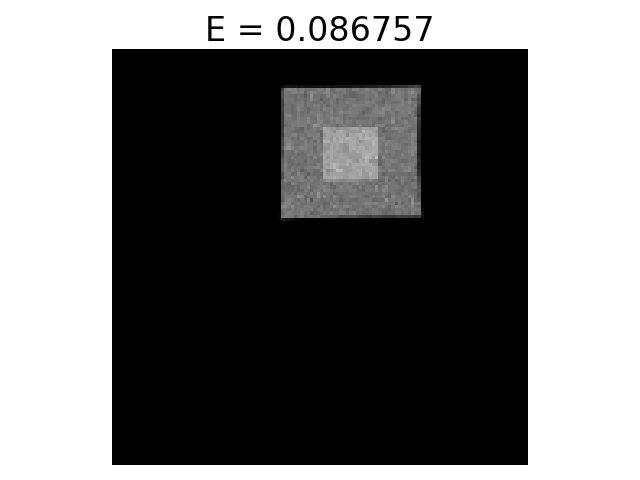} \hspace{-16pt} &
        \includegraphics[width=0.13\textwidth]{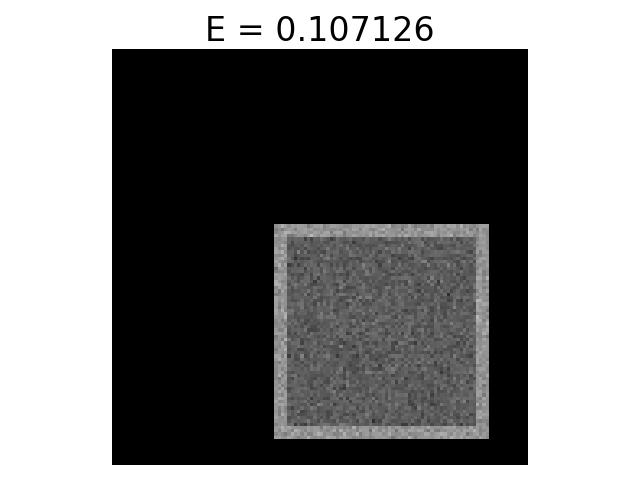} \hspace{-16pt} &
        \includegraphics[width=0.13\textwidth]{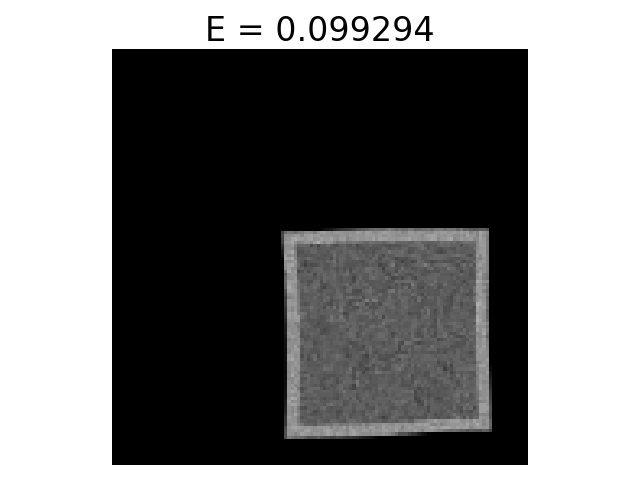} \\

        \textbf{\tiny $ g^\prime \cdot x$} & \textbf{\tiny$(g^\prime \cdot x)_c$} \hspace{-16pt} & \textbf{\tiny $ g^\prime \cdot x$} \hspace{-16pt} & \textbf{\tiny$(g^\prime \cdot x)_c $} \hspace{-16pt} &  \textbf{\tiny $ g^\prime \cdot x$} \hspace{-16pt} & \textbf{\tiny$(g^\prime \cdot x)_c$} \hspace{-16pt} & \textbf{ \tiny$ g^\prime \cdot x $} \hspace{-16pt} & \textbf{\tiny$(g^\prime \cdot x)_c$} \hspace{-6pt} \\

        \hspace{-16pt}
        \includegraphics[width=0.13\textwidth]{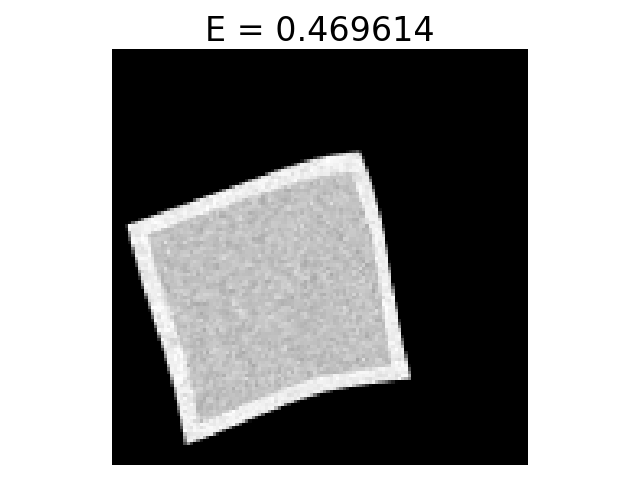} \hspace{-16pt} &
        \includegraphics[width=0.13\textwidth]{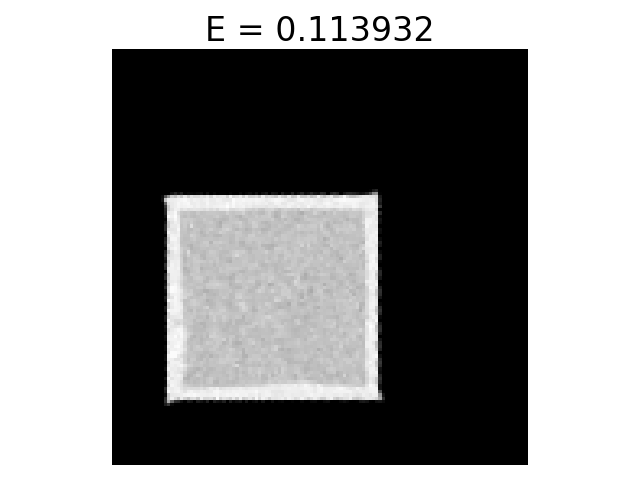} \hspace{-16pt} &
        \includegraphics[width=0.13\textwidth]{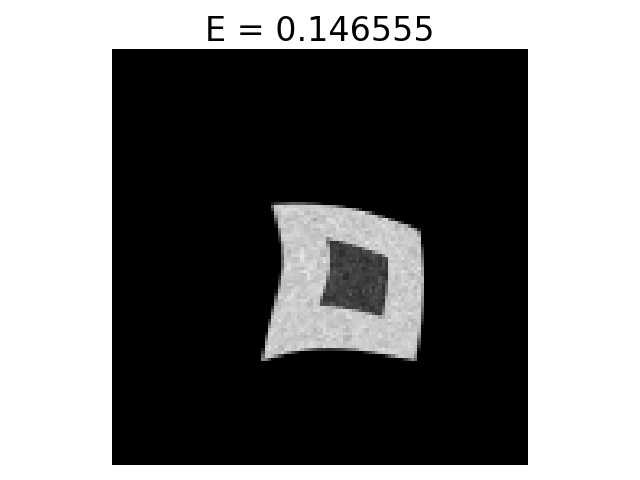} \hspace{-16pt} &
        \includegraphics[width=0.13\textwidth]{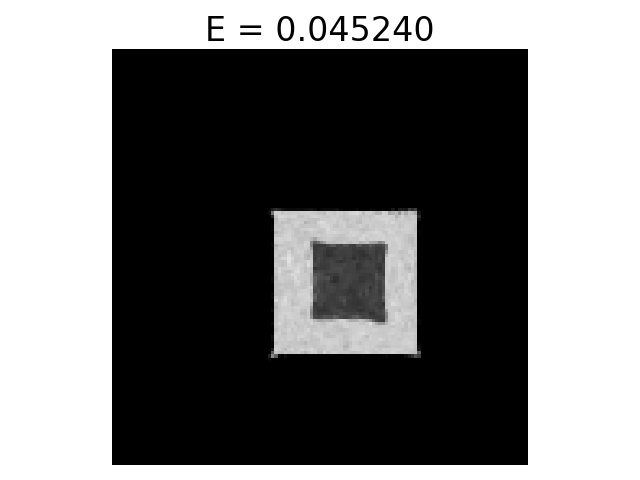} \hspace{-16pt} & \includegraphics[width=0.13\textwidth]{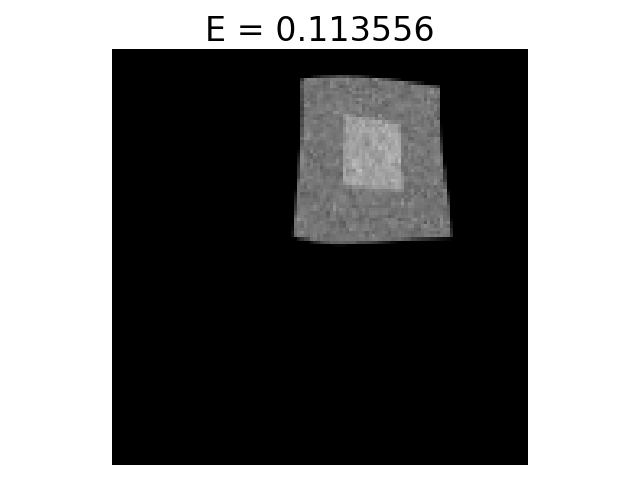} \hspace{-16pt} &
        \includegraphics[width=0.13\textwidth]{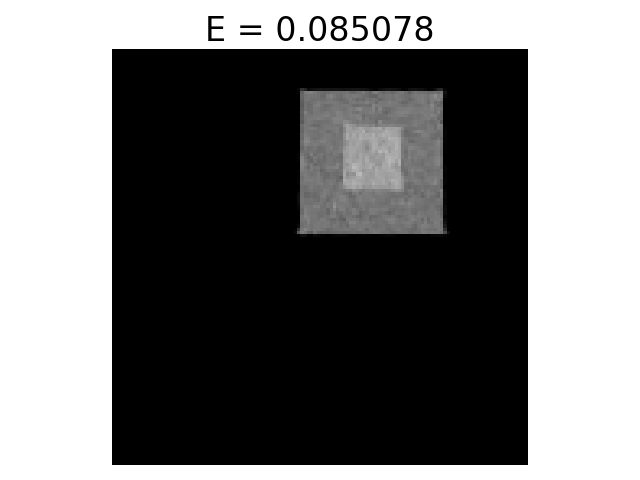} \hspace{-16pt} &
        \includegraphics[width=0.13\textwidth]{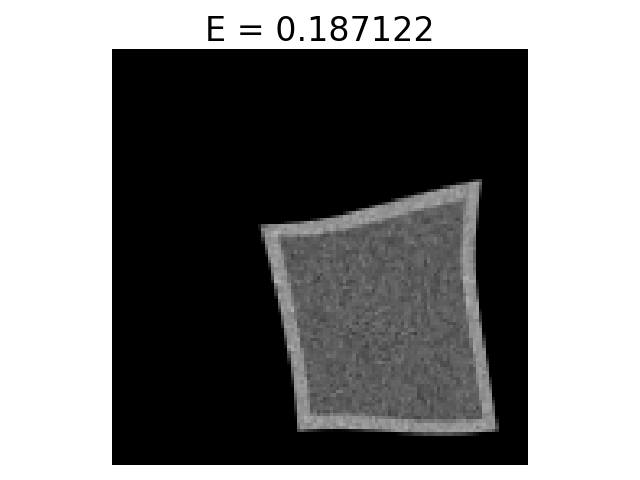} \hspace{-16pt} &
        \includegraphics[width=0.13\textwidth]{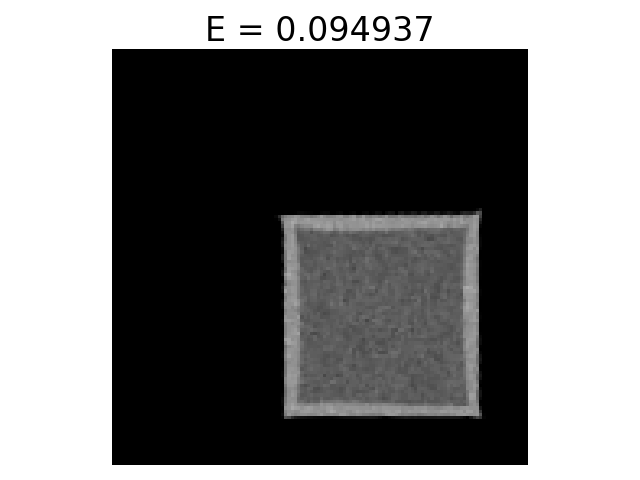} \\
    \end{tabular}
    };
        \draw[line width=1pt] 
            ([xshift=8.6cm,yshift=-0.2cm]table.north west) -- 
            ([xshift=8.6cm,yshift=-4.5cm]table.north west);

            \draw[line width=1pt] 
            ([xshift=4.2cm,yshift=-0.2cm]table.north west) -- 
            ([xshift=4.2cm,yshift=-4.5cm]table.north west);

            \draw[line width=1pt] 
            ([xshift=12.9cm,yshift=-0.2cm]table.north west) -- 
            ([xshift=12.9cm,yshift=-4.5cm]table.north west);

            \draw[->] (1.7,-1.3) -- node [text width=0.9cm, midway, above=-0.1cm, align=center] {{\tiny can}} (2.2,-1.3);
            \draw[->] (1.7,0.9) -- node [text width=0.9cm, midway, above=-0.1cm, align=center] {{\tiny can}} (2.2,0.9);

            \draw[->] (-2.65,-1.3) -- node [text width=0.9cm, midway, above=-0.1cm, align=center] {{\tiny can}} (-2.15,-1.3);
            \draw[->] (-2.65,0.9) -- node [text width=0.9cm, midway, above=-0.1cm, align=center] {{\tiny can}} (-2.15,0.9);

            \draw[->] (-7,-1.3) -- node [text width=0.9cm, midway, above=-0.1cm, align=center] {{\tiny can}} (-6.5,-1.3);
            \draw[->] (-7,0.9) -- node [text width=0.9cm, midway, above=-0.1cm, align=center] {{\tiny can}} (-6.5,0.9);

            \draw[->] (6.05,-1.3) -- node [text width=0.9cm, midway, above=-0.1cm, align=center] {{\tiny can}} (6.55,-1.3);
            \draw[->] (6.05,0.9) -- node [text width=0.9cm, midway, above=-0.1cm, align=center] {{\tiny can}} (6.55,0.9);
    \end{tikzpicture}
    \caption{
    Examples of images $x \in X_E$ and their canonicalised form $x_c$ \textbf{(row one)} with corresponding transformed images $ g^\prime \cdot x \in X_{TE}$ and their canonicalised form $( g^\prime \cdot x)_c$ \textbf{(row two)} with their canonicalisation energies. The experiment is described in Section~\ref{sec:exp_syn}.
    The energies before the canonicalisation steps are very different.
    In contrast, after canonicalising the inputs, the energies are approximately the same. 
    Furthermore, the canonicalised images look very similar to images from $X_E$. 
    This verifies the effectiveness of the canonicalisation step and its invariance empirically.
    }
    \label{fig:test_inv_ex}
\end{figure*}
\section{Experimental Results}\label{sec:experiments}
We 
consider two different tasks: segmentation, where we desire diffeomorphism \emph{equi}variance, and classification, where we aim for diffeomorphism \emph{in}variance. We adopt data augmentation as our gold-standard baseline. While computationally demanding due to the infinite-dimensional nature of the diffeomorphism group, it allows sufficiently expressive models to learn the symmetry by densely sampling the group orbit. We do not aim to outperform this baseline's accuracy, as it represents a theoretical performance ceiling for a given dataset; rather, we aim to match its performance while providing the formal guarantees of equivariance. 
\subsection{Segmentation} \label{sec:exp_segmentation}
This work focuses on binary segmentation, i.e., the output is a binary mask $y_x: \Omega \to \{0,1\}$. 
For segmentation, diffeomorphism equivariance means that if the input image is deformed by a diffeomorphism $g$, the predicted segmentation output should deform exactly the same way.
Hence, the reverse canonicalisation in \eqref{eq:reverse-canon} is exactly the inverse of the canonicalising element $g_x$, i.e., $\Tilde{f}_\theta (x) = g_x^{-1} \cdot \Tilde{f}_\theta (g_x \cdot x)$.
\subsubsection{Synthetic Dataset} \label{sec:exp_syn}
To visualise the functionality and equivariance of DiffeoNN, we consider a synthetic toy example dataset of nested squares with the task of segmenting the inner square~(Appendix~\ref{app:synthetic_data}).
The training dataset $X_E$ contains images with two nested squares of varying sizes and added Gaussian noise, but crucially, restricted to squares aligned to the coordinate axes. 

For evaluation, we create a diffeomorphically transformed dataset $X_{TE}$ by sampling deformation parameters from a radial basis function parametrisation \cite{Forti16032014} to warp the ``canonical'' images in $X_E$. Examples from the datasets $X_E$ and $X_{TE}$ can be found in Figure~\ref{fig:test_inv_ex} and Appendix~\ref{app:synthetic_dataset}. 

\paragraph{Benchmarking Results}
To evaluate the performance of the proposed method, we compare the results of DiffeoNN, the simple inner U-Net of DiffeoNN (na\"ive approach), and an augmented U-Net that is trained in the same way as the inner U-Net of DiffeoNN but on an augmented dataset that includes ``canonical'' and transformed images.
For more details of the setup and training, see Appendix~\ref{appendix:synth_setup}.

We apply the methods to $100$ images from the test dataset $X_{TE}$, and compute the Dice coefficient, Intersection-over-Union (IoU), and pixel-wise accuracy (Table~\ref{tab:benchmark_table}).
DiffeoNN outperforms the na\"ive approach consistently. 
While on average, DiffeoNN does not surpass the performance of our gold-standard baseline (augmented), it produces more stable and reliable results DiffeoNN with fewer outliers, suggesting that our approach is more robust than the augmented U-Net. 
Moreover, DiffeoNN achieves those results with a significantly smaller training dataset, making it a practical and efficient alternative to the data-intensive augmented U-Net.

\paragraph{Invariance of the Canonicalisation}
\begin{table*}[ht]
\caption{Performance of \textbf{DiffeoNN} compared to data augmentation (\textbf{Aug.}) and the inner network of DiffeoNN without augmentation (\textbf{Na\"ive}) for (a/d) \textbf{segmentation of the inner square} on $100$ images from the \textbf{synthetic dataset} (\Cref{app:synthetic_data}), (b/e) \textbf{lung segmentation} on $60$ diffeomorphically transformed chest X-rays from~\cite{realData}, and (c/f) \textbf{homology clas\-si\-fi\-cation} task on $100$ diffeo\-morphi\-cally trans\-formed MNIST~\cite{mnist}. 
DiffeoNN outperforms the na\"ive approach on average and with less variable and thus more stable results. 
On MNIST (c), it matches the average accuracy of the augmented approach for the classification task.
Although, when focusing on the segmentation DiffeoNN does not surpass the average performance of the augmented gold-standard baseline, DiffeoNN produces smaller outliers or fewer outliers, suggesting that our approach is more robust than the augmented U-Net. 
Furthermore, DiffeoNN has the advantage of only needing to be trained on the simple dataset $X_E$, reducing training time and needed computational resources, and requires no retraining.}

\label{tab:benchmark_table}
\begin{subtable}[t]{0.37\textwidth}
 \subcaption{Mean Intersection-over-Union (IoU), Dice coefficient (Dice), and pixel-wise accuracy (Acc.) on \textbf{synthetic dataset}.}
 \label{subtab:bench_syn}
    \begin{center}
    \begin{small}
      \begin{sc}
        \begin{tabular}{lcccr}
          \toprule
          Model  &   IoU & Dice & Acc. \\
          \midrule
          Na\"ive & 0.9276 & 0.9582 & 0.9962 \\
          DiffeoNN & 0.9571 & 0.9777 & 0.9981 \\
          Aug.  & 0.9770 & 0.9878 & 0.9989 \\
          \bottomrule
        \end{tabular}
      \end{sc}
    \end{small}
  \end{center}
  \vskip -0.1in
\end{subtable}
\hfill
\begin{subtable}[t]{0.37\textwidth}
\subcaption{Mean Intersection-over-Union (IoU), Dice coefficient (Dice), and pixel-wise accuracy (Acc.) on diffeomorphically transformed \textbf{chest X-rays}.}
\label{subtab:lung_benchmark_table}
\begin{center}
    \begin{small}
      \begin{sc}
        \begin{tabular}{lcccr}
          \toprule
          Model  &   IoU & \makecell{ Dice } & Acc. \\
          \midrule
          Na\"ive & 0.8759 & 0.9295 & 0.9729 \\
          DiffeoNN & 0.8842 & 0.9327 & 0.9746 \\
          Aug. & 0.9231 & 0.9594 & 0.9836 \\
          \bottomrule
        \end{tabular}
      \end{sc}
    \end{small}
  \end{center}
  \vskip -0.1in
  \end{subtable}
  \hfill
  \begin{subtable}[t]{0.22\textwidth}
  \subcaption{Mean accuracy (Acc.) on diffeo\-morphi\-cally trans\-formed \textbf{MNIST}.
  }
\label{subtab:mnist_benchmark_table}
\begin{center}
    \begin{small}
      \begin{sc}
        \begin{tabular}{lr}
          \toprule
          Model  &    Acc. \\
          \midrule
          Na\"ive & $0.68$\\
          DiffeoNN & $0.82$ \\
          Aug. & $0.82$ \\
          \bottomrule
        \end{tabular}
      \end{sc}
    \end{small}
  \end{center}
  \vskip -0.1in
  \end{subtable}

    \begin{subfigure}[t]{0.37\textwidth}
        \centering
        \includegraphics[height=5.2cm]{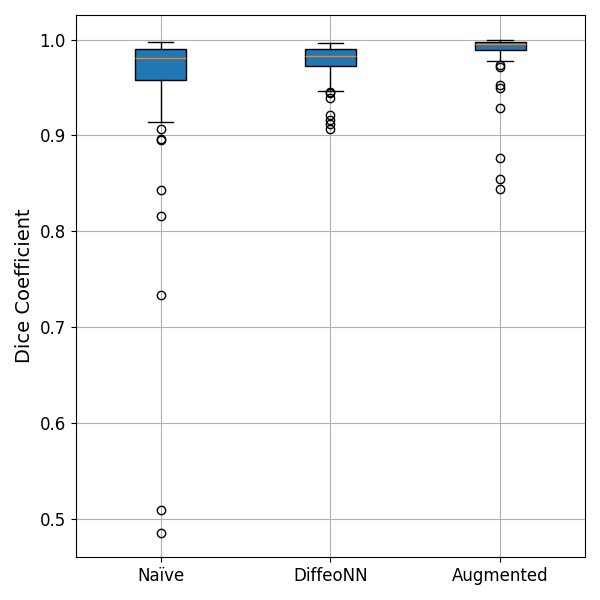}
        \subcaption{Box plots of the Dice coefficient of the \textbf{segmentation} results on the \textbf{synthetic dataset} $X_{TE}$.}
    \end{subfigure}
    \hfill
    \begin{subfigure}[t]{0.37\textwidth}
        \centering
        \includegraphics[height=5.2cm]{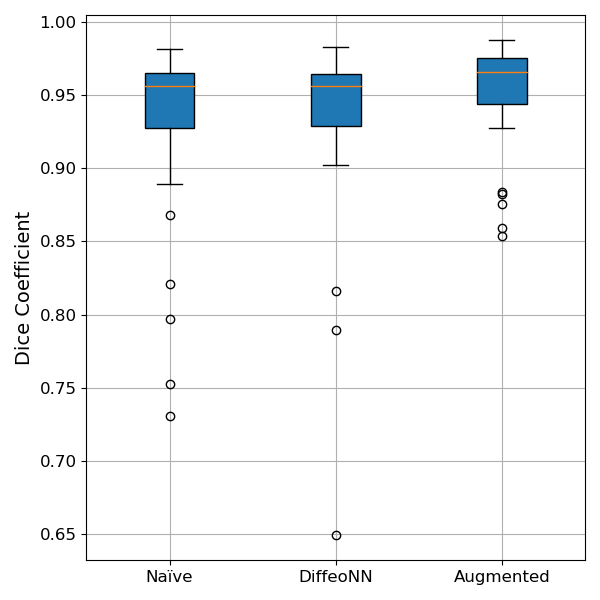}
        \subcaption{Box plots of the Dice coefficient on diffeomorphically transformed \textbf{chest X-rays}.}
    \end{subfigure}
    \hfill
    \begin{subfigure}[t]{0.22\textwidth}
        \centering
        \includegraphics[height=5.2cm]{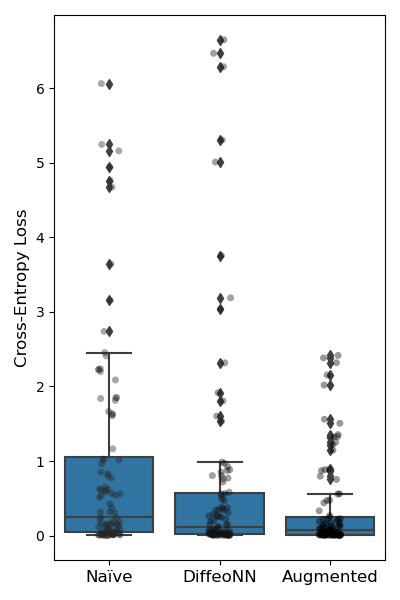}
        \subcaption{Cross-entropy loss on trans\-formed \textbf{MNIST}.}
    \end{subfigure}
\end{table*}
A desirable property of the canonicalisation step is diffeo\-morphism invariance.
As $E_{\text{can}}$ energy is not strictly convex, the canonicalising element $g_x$ is not necessarily unique.
Intuitively, the non-uniqueness is not problematic because as long as the energy $E_{\text{can}}(g_x \cdot x)$ is sufficiently low, the canonicalised input $x_c = g_x \cdot x$ is close enough to the training data $X_E$ for $f_\theta$ to perform well.
This assumption is supported by our empirical observations, while a formal theoretical validation is left for future work.
To verify the approximate invariance in practice, we combine the canonical representations of images in $X_E$ and their randomly transformed counterparts in $X_{TE}$.
We compare the results visually and via their canonicalisation energy, as the solution of the canonicalisation optimisation problem is not unique. 
Visual results can be found in Figure~\ref{fig:test_inv_ex}.
The canonicalised images look very similar to the training dataset $X_E$. 
This is also reflected in the energy levels, as the energies are approximately equal for the canonicalised pairs $(x_c, (g^\prime \cdot x)_c)$.

\subsubsection{Lung Segmentation} \label{sec:exp_lung}

Additionally, we evaluate the performance of DiffeoNN on real-world data, using a dataset with chest X-ray images and their ground-truth lung segmentation from~\cite{realData}.
The original dataset contains images and corresponding ground-truth segmentations into three different classes (``Non-Covid'', ``Covid'', and ``Non-Covid-Pneu\-mo\-nia'').
We combine the images and corresponding ground-truth segmentations of the initial three classes into one dataset, which is then split into a training dataset of $234$ image-segmentation pairs, a validation dataset of $60$ pairs, and a test dataset of $60$ pairs.
We then proceed as in Section~\ref{sec:exp_syn} to create a dataset of diffeomorphically transformed images. 

\paragraph{Benchmarking Results}
We apply DiffeoNN, the inner U-Net (na\"ive approach) and an augmented U-Net, which is trained on the original and the diffeomorphically transformed training dataset, to the diffeomorphically transformed images from the test dataset.
Visual results are presented in Figure~\ref{fig:exp_lung} and further details on the experimental setup, training, and additional examples in Appendix~\ref{appendix:Lung_exp}.
The results closely mirror those obtained on the synthetic dataset, see Table~\ref{subtab:lung_benchmark_table}.
On average, DiffeoNN outperforms the na\"ive approach, validating the effectiveness of our approach. 
The canonicalisation step pushes the input image towards the training dataset by moving the thorax to a more central position and aligning the shoulders, which makes it easier for the inner U-Net to predict an accurate lung segmentation.
While our method performs slightly below the augmented gold-standard baseline, it offers more flexibility by not relying on data augmentation or retraining, making it a practical and robust solution in real-world applications.

\subsection{Classification} \label{sec:exp_classification}
To illustrate the ability of our approach to achieve invariance, i.e., $\Tilde{f}_\theta (x) = \Tilde{f}_\theta (g \cdot x)$, we test it on the MNIST dataset~\cite{mnist} for the task of determining the genus, i.e., counting topological holes. 
The typical classification task of identifying the written digit is not diffeomorphism invariant as, e.g., ``6'' can be mapped to ``9'' by a diffeomorphism, so they would be assigned to the same class.
Similarly to Section~\ref{sec:exp_segmentation}, we transform the MNIST dataset $X_E$ diffeomorphically to create a second dataset $X_{TE}$ for testing.

\paragraph{Benchmarking Results}
We compare the performance of DiffeoNN to the na\"ive approach, i.e., the inner convolutional classifier for determining the genus, and an augmented approach. 
The latter is trained in analogy with other experiments, on images from the original and transformed MNIST. 
The mean accuracy over $100$ transformed MNIST images can be found in Table~\ref{subtab:mnist_benchmark_table} and visual examples in Appendix~\ref{appendix:mnist}.
Our canonicalisation strategy noticeably improves the na\"ive approach, matching the performance of the augmented classifier, yet without retraining on the augmented data.

\section{Conclusion}

By combining Lie group theory with differentiable image registration, we have developed a robust energy-based canonicalisation strategy for diffeomorphism equivariance. 
Our approach is highly data-efficient, achieving approximate equivariance using only a small dataset without augmentation. The primary advantage of this approach is its modularity: it can render any pretrained network diffeomorphism-equivariant without altering the inner task-performing model.

Crucially, we have extended the applicability of energy canonicalisation by demonstrating its robustness and effectiveness even when applied to groups as large as the group of diffeomorphisms. Our results highlight a significant gap in the broader literature: despite accumulating empirical evidence of efficacy, there are few theoretical results regarding the stability and generalisability of canonicalisation. A vital direction for future research remains to characterise learning bounds and sufficiency of approximate canonicalisation for achieving full model equivariance.

\newpage
\section*{Impact Statement}
The work presented here is intended to advance the fields of Machine Learning and Geometric Deep Learning. 
We do not expect our work to have any specifically notable potential societal consequences apart from those usually associated with such advancements.

\bibliographystyle{icml2026}
\bibliography{references}

\newpage
\appendix
\crefalias{section}{appendix}
\onecolumn
\raggedbottom

\section{Synthetic Dataset}\label{app:synthetic_data}
\subsection{Examples of Synthetic Dataset} \label{app:synthetic_dataset}
\begin{figure}[ht]
    \centering
    \begin{tabular}{cccc|cccc}
        \vspace{0.5em}
        \textbf{$X_{E}$} & \textbf{$Y_{E}$} & \textbf{$X_{TE}$} & \textbf{$Y_{TE}$} & \textbf{$X_{E}$} & \textbf{$Y_{E}$} & \textbf{$X_{TE}$} & \textbf{$Y_{TE}$} \\
        
        \includegraphics[width=0.10\textwidth]{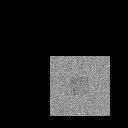} &
        \includegraphics[width=0.10\textwidth]{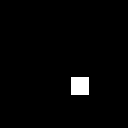} &
        \includegraphics[width=0.10\textwidth]{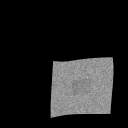} &
        \includegraphics[width=0.10\textwidth]{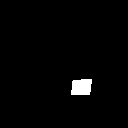} & \includegraphics[width=0.10\textwidth]{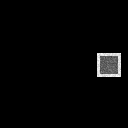} &
        \includegraphics[width=0.10\textwidth]{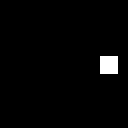} &
        \includegraphics[width=0.10\textwidth]{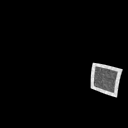} &
        \includegraphics[width=0.10\textwidth]{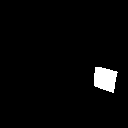} \\

        \includegraphics[width=0.10\textwidth]{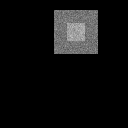} &
        \includegraphics[width=0.10\textwidth]{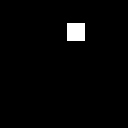} &
        \includegraphics[width=0.10\textwidth]{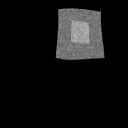} &
        \includegraphics[width=0.10\textwidth]{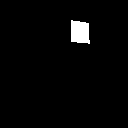} & \includegraphics[width=0.10\textwidth]{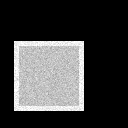} &
        \includegraphics[width=0.10\textwidth]{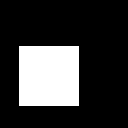} &
        \includegraphics[width=0.10\textwidth]{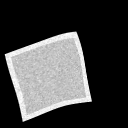} &
        \includegraphics[width=0.10\textwidth]{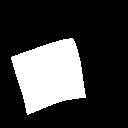} \\

        \includegraphics[width=0.10\textwidth]{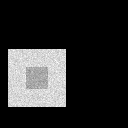} &
        \includegraphics[width=0.10\textwidth]{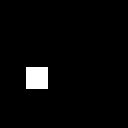} &
        \includegraphics[width=0.10\textwidth]{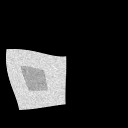} &
        \includegraphics[width=0.10\textwidth]{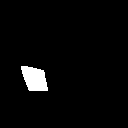} & \includegraphics[width=0.10\textwidth]{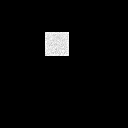} &
        \includegraphics[width=0.10\textwidth]{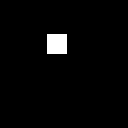} &
        \includegraphics[width=0.10\textwidth]{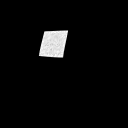} &
        \includegraphics[width=0.10\textwidth]{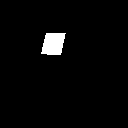} 
    \end{tabular}
    \caption{Examples of the generated square images $X_{E}$ \textbf{(column one)} and their binary segmentation $Y_{E}$ \textbf{(column two)}, as well as a random diffeomorphic transformations of those images ,$X_{TE}$, \textbf{(column three)} and the corresponding output $Y_{TE}$ \textbf{(column four)}.
    The images in $X_{E}$ contain two nested squares of varying size and colour, where the squares' edges are parallel to the image edges with added Gaussian noise.
    The binary segmentation $Y_{E}$ indicates the inner square.
    The images $X_{TE}$ and their corresponding segmentations $Y_{TE}$ are obtained by transforming the images $X_{E}$ and their binary segmentations $Y_{E}$ with the same randomly chosen transformations.}
\end{figure}
\subsection{Experiments on the Synthetic Dataset} \label{appendix:synth_setup}
For the inner network, we train a simple U-Net~\cite{unet} with a learning rate of $5\cdot 10^{-5}$ and a batch size of $2$ for $10$ epochs on the training dataset $X_E$ using Adam~\cite{adam}.

For the image similarity energy (see Section~\ref{sec:can_energy}) a VAE-CNN with a ten-dimensional latent space is trained on $X_E$ using Adam~\cite{adam} with a batch size of $1$, a learning rate of $5\cdot 10^{-5}$ and $50$ epochs, following the implementation of~\cite{vae, ZaksLieLAC}. 
The implementation for the convex adversarial discriminator closely follows \cite{advReg1, advReg2, ZaksLieLAC}. 
The images for are transformed on-the-fly with randomly chosen diffeomorphisms similar to the ones, used for creating $X_{TE}$.
The gradient penalty weight is set to $\mu=10$ and the learning rate to $1 \cdot 10^{-4}$, the batch size to~$16$, and the number of epochs to~$50$. 
The used optimiser is the Adam optimiser~\cite{adam}.

We set the scalar weights in $E_{\text{can}}$, see \eqref{eq:E_XE} and \eqref{eq:E_reg}, to $\lambda_{\text{adv}}= 0.01$, $\lambda_{\text{VAE}} = 1 \cdot 10^{-5}$,  $\lambda_{\Delta} = 1$, and $\lambda_{\mathcal{J}} = 10$.
\begin{figure}[H]
    \centering
    \begin{tabular}{ccccc}
        \textbf{\makecell{Input $x$}} & \textbf{\makecell{Canonicalised \\ Input $x_c$}} & \textbf{\makecell{Segmentation \\ of $x_c$}} & \textbf{\makecell{Output \\ DiffeoNN}} & \textbf{\makecell{Ground-Truth \\ Segmentation}} \\
        \includegraphics[width=0.17\textwidth]{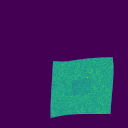} &
        \includegraphics[width=0.17\textwidth]{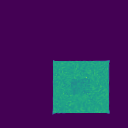} &
        \includegraphics[width=0.17\textwidth]{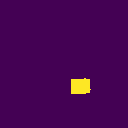} &
        \includegraphics[width=0.17\textwidth]{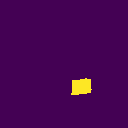} &
        \includegraphics[width=0.17\textwidth]{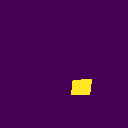} \\

        \includegraphics[width=0.17\textwidth]{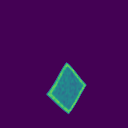} &
        \includegraphics[width=0.17\textwidth]{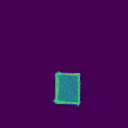} &
        \includegraphics[width=0.17\textwidth]{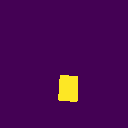} &
        \includegraphics[width=0.17\textwidth]{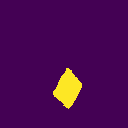} &
        \includegraphics[width=0.17\textwidth]{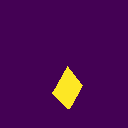} \\

        \includegraphics[width=0.17\textwidth]{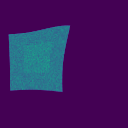} &
        \includegraphics[width=0.17\textwidth]{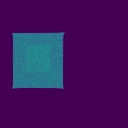} &
        \includegraphics[width=0.17\textwidth]{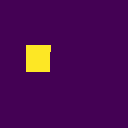} &
        \includegraphics[width=0.17\textwidth]{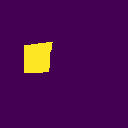} &
        \includegraphics[width=0.17\textwidth]{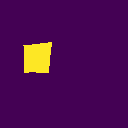} \\

        \includegraphics[width=0.17\textwidth]{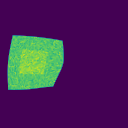} &
        \includegraphics[width=0.17\textwidth]{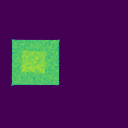} &
        \includegraphics[width=0.17\textwidth]{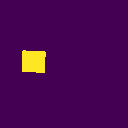} &
        \includegraphics[width=0.17\textwidth]{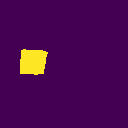} &
        \includegraphics[width=0.17\textwidth]{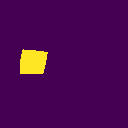} \\
    
        \includegraphics[width=0.17\textwidth]{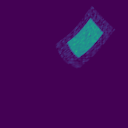} &
        \includegraphics[width=0.17\textwidth]{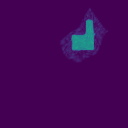} &
        \includegraphics[width=0.17\textwidth]{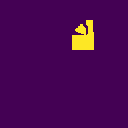} &
        \includegraphics[width=0.17\textwidth]{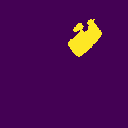} &
        \includegraphics[width=0.17\textwidth]{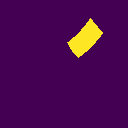} \\

        \includegraphics[width=0.17\textwidth]{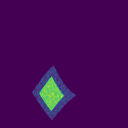} &
        \includegraphics[width=0.17\textwidth]{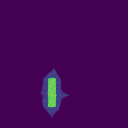} &
        \includegraphics[width=0.17\textwidth]{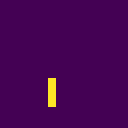} &
        \includegraphics[width=0.17\textwidth]{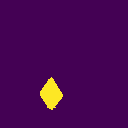} &
        \includegraphics[width=0.17\textwidth]{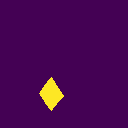} \\
        
    \end{tabular}
    \caption{Examples of results of DiffeoNN step by step on the synthetic test dataset $X_{TE}$.
    The \textbf{first four rows} show a nearly perfect segmentation through DiffeoNN. 
    The canonicalised input images look very similar to the training dataset of squares $X_E$, and the output segmentation is very similar to the ground-truth segmentation.
    \textbf{Row five} shows an example where the canonicalisation step failed. 
    As a result, the segmentation of the canonicalised image $x_c$ contains artifacts. 
    The artifacts are transferred to the output segmentation of the input $x$, which differs noticeably from the ground-truth segmentation. 
    In \textbf{row six}, we show an example where the canonicalisation step leads to an image $x_c$ that contains rectangles rather than squares. Even though this image does not look exactly like the training images with squares $X_E$, the segmentation output for $x_c$ is still very accurate, which leads to a segmentation output for $x$ that is nearly identical to the ground truth.}
\end{figure}
\begin{figure}[H]
    \centering
    \begin{tabular}{ccccc}
    \textbf{\makecell{Input}} &
    \textbf{\makecell{Output \\ Na\"ive U-Net}} &
    \textbf{\makecell{Output \\ DiffeoNN}} &
    \textbf{\makecell{Output \\ Augmented \\ U-Net}} &
    \textbf{\makecell{Ground Truth}} \\

    \includegraphics[width=0.17\textwidth]{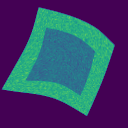} &
    \includegraphics[width=0.17\textwidth]{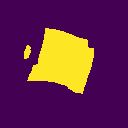} &
    \includegraphics[width=0.17\textwidth]{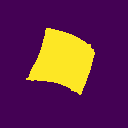} &
    \includegraphics[width=0.17\textwidth]{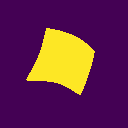} &
    \includegraphics[width=0.17\textwidth]{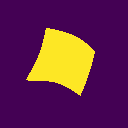} \\

    \includegraphics[width=0.17\textwidth]{graphics/diffeo_res/11_moving.png} &
    \includegraphics[width=0.17\textwidth]{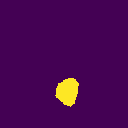} &
    \includegraphics[width=0.17\textwidth]{graphics/diffeo_res/11_pred_orig_mask.png} &
    \includegraphics[width=0.17\textwidth]{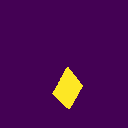} &
    \includegraphics[width=0.17\textwidth]{graphics/aug_res/11_mask.png} \\

    \includegraphics[width=0.17\textwidth]{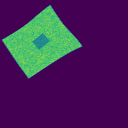} &
    \includegraphics[width=0.17\textwidth]{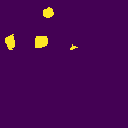} &
    \includegraphics[width=0.17\textwidth]{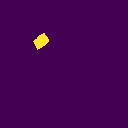} &
    \includegraphics[width=0.17\textwidth]{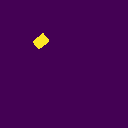} &
    \includegraphics[width=0.17\textwidth]{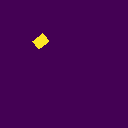} \\

    \includegraphics[width=0.17\textwidth]{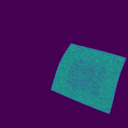} &
    \includegraphics[width=0.17\textwidth]{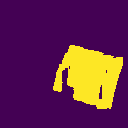} &
    \includegraphics[width=0.17\textwidth]{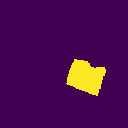} &
    \includegraphics[width=0.17\textwidth]{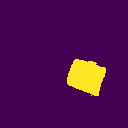} &
    \includegraphics[width=0.17\textwidth]{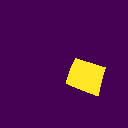} \\

    \includegraphics[width=0.17\textwidth]{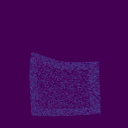} &
    \includegraphics[width=0.17\textwidth]{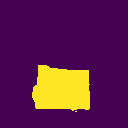} &
    \includegraphics[width=0.17\textwidth]{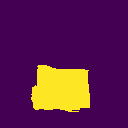} &
    \includegraphics[width=0.17\textwidth]{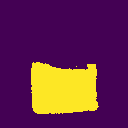} &
    \includegraphics[width=0.17\textwidth]{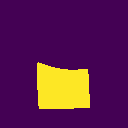} \\

    \includegraphics[width=0.17\textwidth]{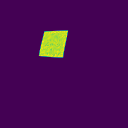} &
    \includegraphics[width=0.17\textwidth]{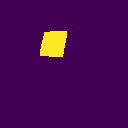} &
    \includegraphics[width=0.17\textwidth]{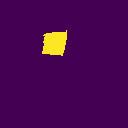} &
    \includegraphics[width=0.17\textwidth]{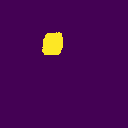} &
    \includegraphics[width=0.17\textwidth]{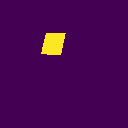} \\
    \end{tabular}
    \caption{Examples comparing the results of the na\"ive U-Net, DiffeoNN, and the augmented U-Net on $ X_{TE}$ (synthetic dataset).
    \textbf{Column one} shows the input image from $x \in X_{TE}$, and \textbf{column five} shows the corresponding ground-truth segmentation.
In \textbf{column two to four}, we present the segmentation output of $x$ by different networks.
In the \textbf{first four rows}, the segmentation of the na\"ive U-Net is noticeably different from the ground truth, while the segmentations of DiffeoNN and the augmented U-Net are similar.
The \textbf{last two rows} show examples where the augmented U-Net performs worse than the other two methods.
}

\end{figure}

\section{Visual Results: Lung Segmentation}\label{appendix:Lung_exp}
We train the inner U-Net as well as the adversarial network and VAE on the original chest X-rays similar to Section~\ref{appendix:synth_setup}.
We change the latent space dimension of the VAE to $100$ and perform $100$ gradient steps for finding a canonicalising element.
The weights in the canonicalisation energy $E_{\text{can}}$ 
are set to $\lambda_{\text{adv}}= 0.001$, $\lambda_{\text{VAE}} = 0.001$,  $\lambda_{\Delta} = 1$, and $\lambda_{\mathcal{J}} = 10$.
\begin{figure}[H]
    \centering
    \begin{tabular}{c|cc|cccc}
        \textbf{\footnotesize \makecell{Input $x$}} & \textbf{\footnotesize \makecell{Canonicalised \\ Input $x_c$}} & \textbf{\footnotesize \makecell{Segmentation \\ of $x_c$}} & \textbf{\footnotesize \makecell{Output \\ DiffeoNN}} & \textbf{\footnotesize \makecell{Output \\ Na\"ive U-Net}} & \textbf{\footnotesize \makecell{Output \\ Augmented \\ U-Net}} & \textbf{\footnotesize \makecell{Ground-Truth \\ Segmentation}} \\
    
        \includegraphics[width=0.12\textwidth]{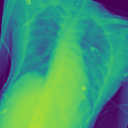} &
        \includegraphics[width=0.12\textwidth]{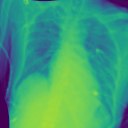} &
        \includegraphics[width=0.12\textwidth]{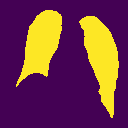} &
        \includegraphics[width=0.12\textwidth]{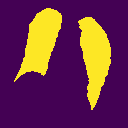} &
        \includegraphics[width=0.12\textwidth]{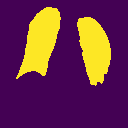} &
        \includegraphics[width=0.12\textwidth]{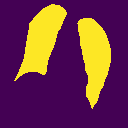} &
        \includegraphics[width=0.12\textwidth]{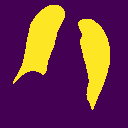} \\

        \includegraphics[width=0.12\textwidth]{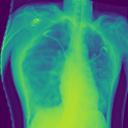} &
        \includegraphics[width=0.12\textwidth]{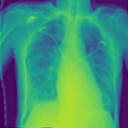} &
        \includegraphics[width=0.12\textwidth]{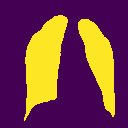} &
        \includegraphics[width=0.12\textwidth]{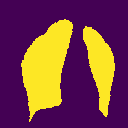} &
        \includegraphics[width=0.12\textwidth]{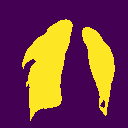} &
        \includegraphics[width=0.12\textwidth]{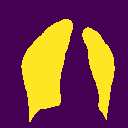} &
        \includegraphics[width=0.12\textwidth]{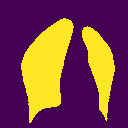} \\

        \includegraphics[width=0.12\textwidth]{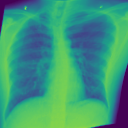} &
        \includegraphics[width=0.12\textwidth]{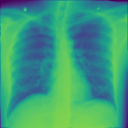} &
        \includegraphics[width=0.12\textwidth]{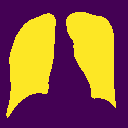} &
        \includegraphics[width=0.12\textwidth]{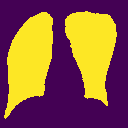} &
        \includegraphics[width=0.12\textwidth]{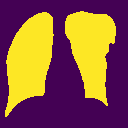} &
        \includegraphics[width=0.12\textwidth]{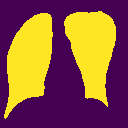} &
        \includegraphics[width=0.12\textwidth]{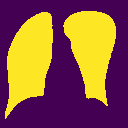} \\

        \includegraphics[width=0.12\textwidth]{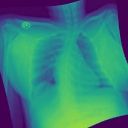} &
        \includegraphics[width=0.12\textwidth]{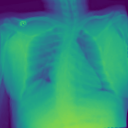} &
        \includegraphics[width=0.12\textwidth]{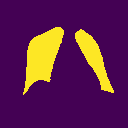} &
        \includegraphics[width=0.12\textwidth]{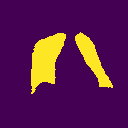} &
        \includegraphics[width=0.12\textwidth]{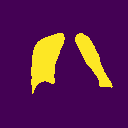} &
        \includegraphics[width=0.12\textwidth]{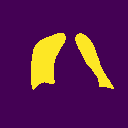} &
        \includegraphics[width=0.12\textwidth]{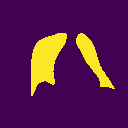} \\

        \includegraphics[width=0.12\textwidth]{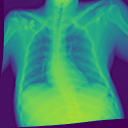} &
        \includegraphics[width=0.12\textwidth]{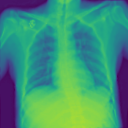} &
        \includegraphics[width=0.12\textwidth]{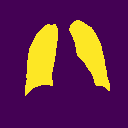} &
        \includegraphics[width=0.12\textwidth]{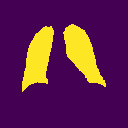} &
        \includegraphics[width=0.12\textwidth]{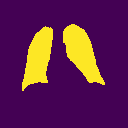} &
        \includegraphics[width=0.12\textwidth]{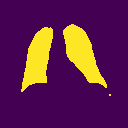} &
        \includegraphics[width=0.12\textwidth]{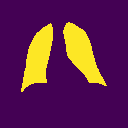} \\

        \includegraphics[width=0.12\textwidth]{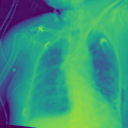} &
        \includegraphics[width=0.12\textwidth]{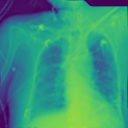} &
        \includegraphics[width=0.12\textwidth]{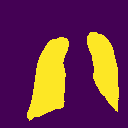} &
        \includegraphics[width=0.12\textwidth]{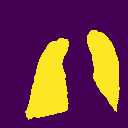} &
        \includegraphics[width=0.12\textwidth]{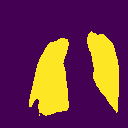} &
        \includegraphics[width=0.12\textwidth]{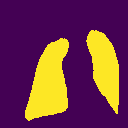} &
        \includegraphics[width=0.12\textwidth]{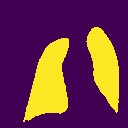} \\

        \includegraphics[width=0.12\textwidth]{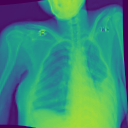} &
        \includegraphics[width=0.12\textwidth]{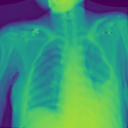} &
        \includegraphics[width=0.12\textwidth]{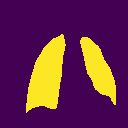} &
        \includegraphics[width=0.12\textwidth]{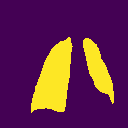} &
        \includegraphics[width=0.12\textwidth]{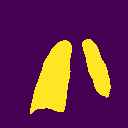} &
        \includegraphics[width=0.12\textwidth]{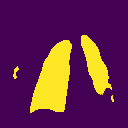} &
        \includegraphics[width=0.12\textwidth]{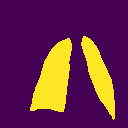} 

    \end{tabular}
    \caption{Lung segmentation of diffeomorphically transformed chest X-ray images from~\cite{realData}.
    Shown are the steps of DiffeoNN \textbf{(column two and three)}, as well as the final outputs of DiffeoNN, the inner U-Net of DiffeoNN without augmentation (na\"ive), and an augmented U-Net \textbf{(column four, five, and six).}
    In the canonicalised images, the thorax was moved to a more central position and the shoulders were aligned, which makes it easier for the inner U-Net to predict an accurate lung segmentation.
    If the canonicalisation step works, DiffeoNN produces more accurate lung segmentations than the na\"ive approach.
    The segmentation output of DiffeoNN is very similar to the output of the augmented U-Net (and the ground truth) and does occasionally even provide a more accurate segmentation (see last row).}
\end{figure}
\begin{figure}[H]
    \centering
    \begin{tabular}{c|cc|cccc}
        \textbf{\footnotesize \makecell{Input $x$}} & \textbf{\footnotesize \makecell{Canonicalised \\ Input $x_c$}} & \textbf{\footnotesize \makecell{Segmentation \\ of $x_c$}} & \textbf{\footnotesize \makecell{Output \\ DiffeoNN}} & \textbf{\footnotesize \makecell{Output \\ Na\"ive U-Net}} & \textbf{\footnotesize \makecell{Output \\ Augmented \\ U-Net}} & \textbf{\footnotesize \makecell{Ground-Truth \\ Segmentation}} \\
        
        \includegraphics[width=0.10\textwidth]{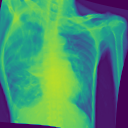} &
        \includegraphics[width=0.10\textwidth]{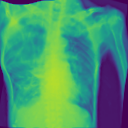} &
        \includegraphics[width=0.10\textwidth]{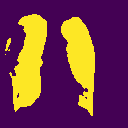} &
        \includegraphics[width=0.10\textwidth]{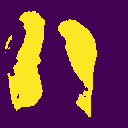} &
        \includegraphics[width=0.10\textwidth]{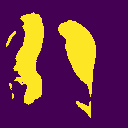} &
        \includegraphics[width=0.10\textwidth]{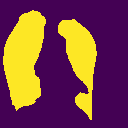} &
        \includegraphics[width=0.10\textwidth]{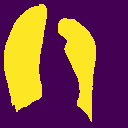} \\

        \includegraphics[width=0.10\textwidth]{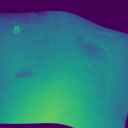} &
        \includegraphics[width=0.10\textwidth]{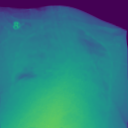} &
        \includegraphics[width=0.10\textwidth]{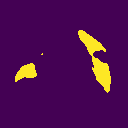} &
        \includegraphics[width=0.10\textwidth]{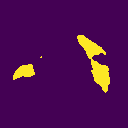} &
        \includegraphics[width=0.10\textwidth]{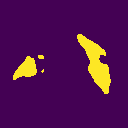} &
        \includegraphics[width=0.10\textwidth]{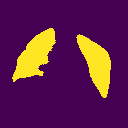} &
        \includegraphics[width=0.10\textwidth]{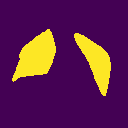} \\

        \includegraphics[width=0.10\textwidth]{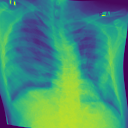} &
        \includegraphics[width=0.10\textwidth]{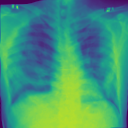} &
        \includegraphics[width=0.10\textwidth]{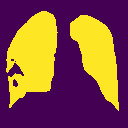} &
        \includegraphics[width=0.10\textwidth]{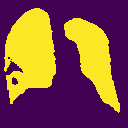} &
        \includegraphics[width=0.10\textwidth]{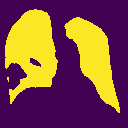} &
        \includegraphics[width=0.10\textwidth]{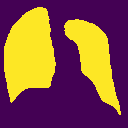} &
        \includegraphics[width=0.10\textwidth]{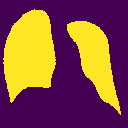} \\

        \includegraphics[width=0.10\textwidth]{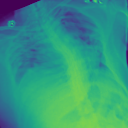} &
        \includegraphics[width=0.10\textwidth]{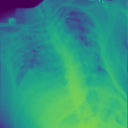} &
        \includegraphics[width=0.10\textwidth]{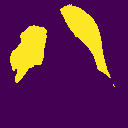} &
        \includegraphics[width=0.10\textwidth]{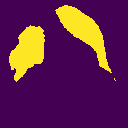} &
        \includegraphics[width=0.10\textwidth]{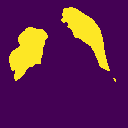} &
        \includegraphics[width=0.10\textwidth]{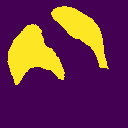} &
        \includegraphics[width=0.10\textwidth]{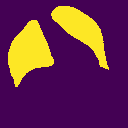} \\

        \includegraphics[width=0.10\textwidth]{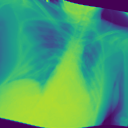} &
        \includegraphics[width=0.10\textwidth]{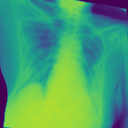} &
        \includegraphics[width=0.10\textwidth]{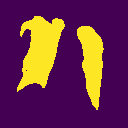} &
        \includegraphics[width=0.10\textwidth]{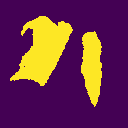} &
        \includegraphics[width=0.10\textwidth]{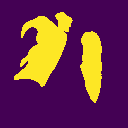} &
        \includegraphics[width=0.10\textwidth]{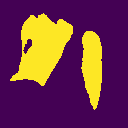} &
        \includegraphics[width=0.10\textwidth]{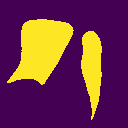} \\

    \end{tabular}
    \caption{Lung segmentation of diffeomorphically transformed chest X-ray images from~\cite{realData}, where the results of the naive approach and DiffeoNN are inaccurate. The segmentations show visible gaps or are missing parts.
    Either two inner U-Net could not segment parts of the image properly, even after the canonicalisation step and/or the canonicalisation step did not work properly.
    The segmentation of DiffeoNN remains slightly more accurate than the na\"ive one.
    As a rule, the augmented U-Net does provide a more accurate segmentation in those cases.
    }
\end{figure}
\begin{figure}[H]
    \centering
    \begin{tabular}{c|cc|cccc}
        \textbf{\footnotesize \makecell{Input $x$}} & \textbf{\footnotesize \makecell{Canonicalised \\ Input $x_c$}} & \textbf{\footnotesize \makecell{Segmentation \\ of $x_c$}} & \textbf{\footnotesize \makecell{Output \\ DiffeoNN}} & \textbf{\footnotesize \makecell{Output \\ Na\"ive U-Net}} & \textbf{\footnotesize \makecell{Output \\ Augmented \\ U-Net}} & \textbf{\footnotesize \makecell{Ground-Truth \\ Segmentation}} \\
        
        \includegraphics[width=0.10\textwidth]{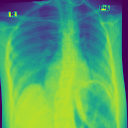} &
        \includegraphics[width=0.10\textwidth]{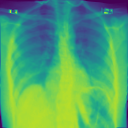} &
        \includegraphics[width=0.10\textwidth]{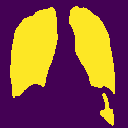} &
        \includegraphics[width=0.10\textwidth]{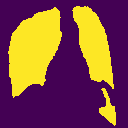} &
        \includegraphics[width=0.10\textwidth]{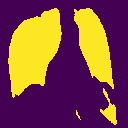} &
        \includegraphics[width=0.10\textwidth]{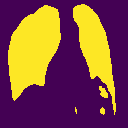} &
        \includegraphics[width=0.10\textwidth]{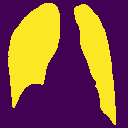} \\
        
        \includegraphics[width=0.10\textwidth]{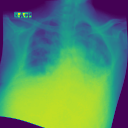} &
        \includegraphics[width=0.10\textwidth]{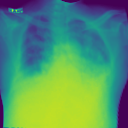} &
        \includegraphics[width=0.10\textwidth]{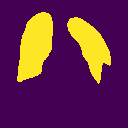} &
        \includegraphics[width=0.10\textwidth]{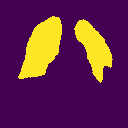} &
        \includegraphics[width=0.10\textwidth]{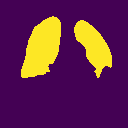} &
        \includegraphics[width=0.10\textwidth]{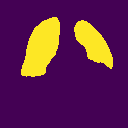} &
        \includegraphics[width=0.10\textwidth]{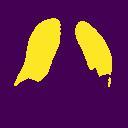} \\

        \includegraphics[width=0.10\textwidth]{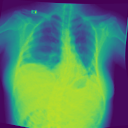} &
        \includegraphics[width=0.10\textwidth]{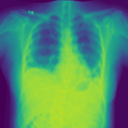} &
        \includegraphics[width=0.10\textwidth]{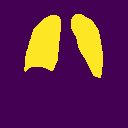} &
        \includegraphics[width=0.10\textwidth]{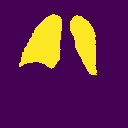} &
        \includegraphics[width=0.10\textwidth]{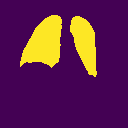} &
        \includegraphics[width=0.10\textwidth]{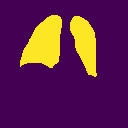} &
        \includegraphics[width=0.10\textwidth]{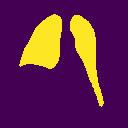} \\

    \end{tabular}
    \caption{Lung segmentation of diffeomorphically transformed chest X-ray images from~\cite{realData}, where the data (chest X-ray and ground-truth segmentation) are of a variability that could not be learned by either of the three methods. 
    The canonicalised images do look very close to the training dataset $X_E$, but the final segmentation output of DiffeoNN does not match the ground-truth segmentation. 
    Nevertheless neither does the output of the na\"ive and the augmented approach.
    }
\end{figure}
\section{Results: Homology Classification on MNIST}\label{appendix:mnist}
\begin{figure}[H]
    \centering
    \begin{tabular}{cccc|cccc}
        \textbf{\makecell{Output \\ Na\"ive}} & 
         \textbf{\makecell{Output \\ Aug. }} & \textbf{\makecell{Output \\ DiffeoNN}} & \textbf{\makecell{Ground \\ Truth}}  & 
        \textbf{\makecell{Output \\ Na\"ive}} & 
         \textbf{\makecell{Output \\ Aug. }} & \textbf{\makecell{Output \\ DiffeoNN}} & \textbf{\makecell{Ground \\ Truth}} \\

        \includegraphics[width=0.10\textwidth]{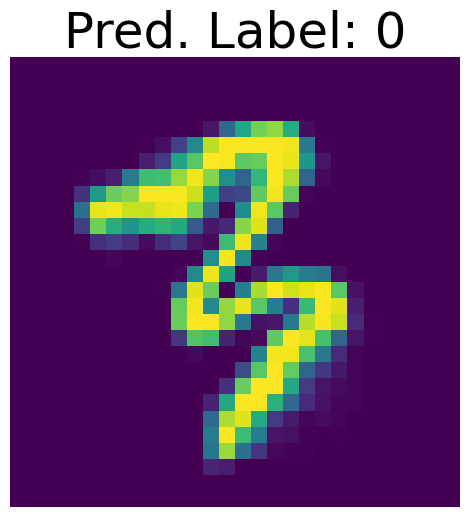} &
        \includegraphics[width=0.10\textwidth]{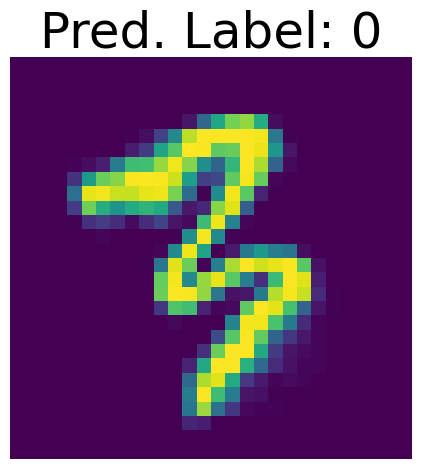} &
        \includegraphics[width=0.10\textwidth]{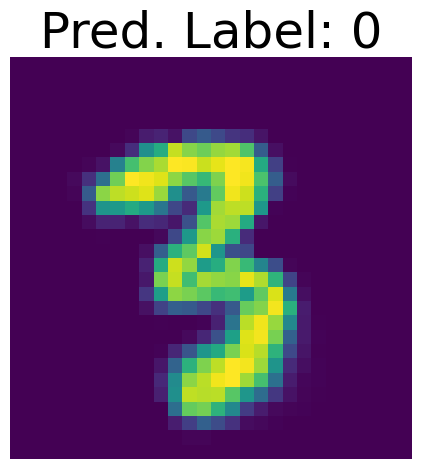} &
        \includegraphics[width=0.10\textwidth]{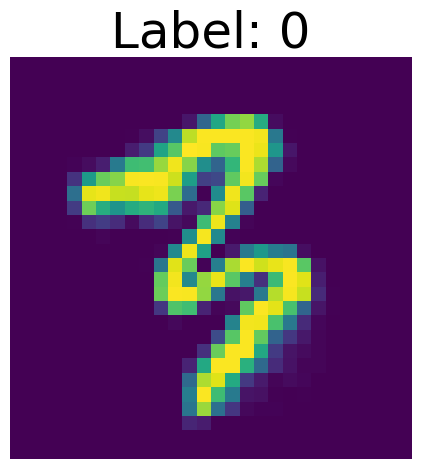} &
        \includegraphics[width=0.10\textwidth]{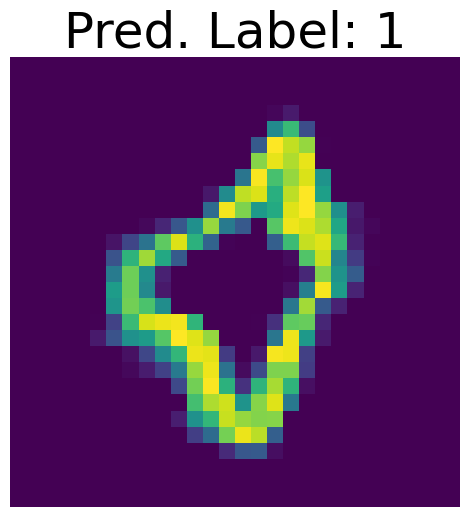} &
        \includegraphics[width=0.10\textwidth]{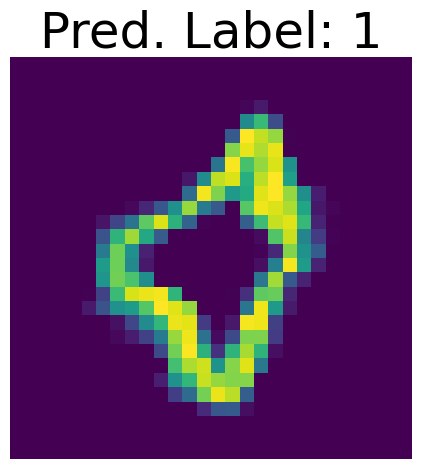} &
        \includegraphics[width=0.10\textwidth]{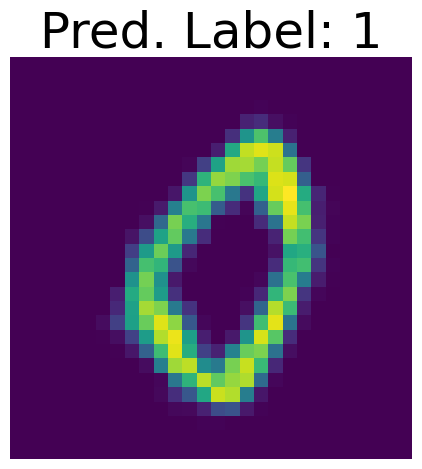} &
        \includegraphics[width=0.10\textwidth]{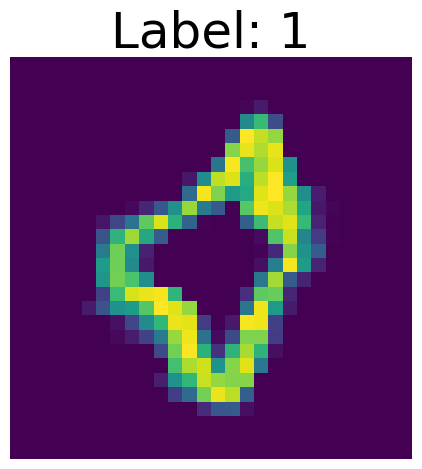}  \\ 

        \includegraphics[width=0.10\textwidth]{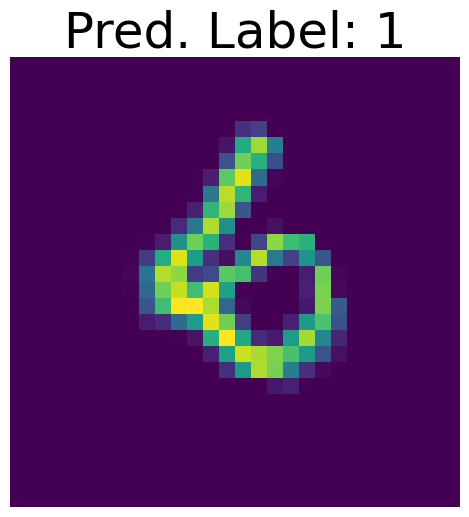} &
        \includegraphics[width=0.10\textwidth]{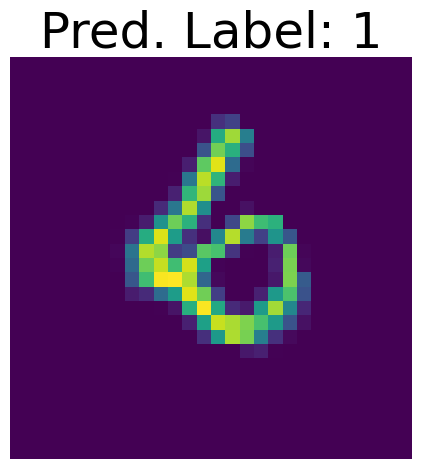} &
        \includegraphics[width=0.10\textwidth]{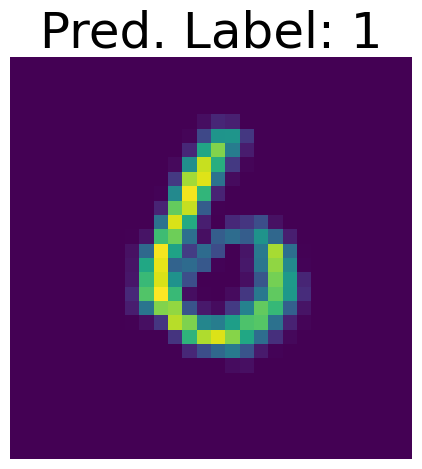} &
        \includegraphics[width=0.10\textwidth]{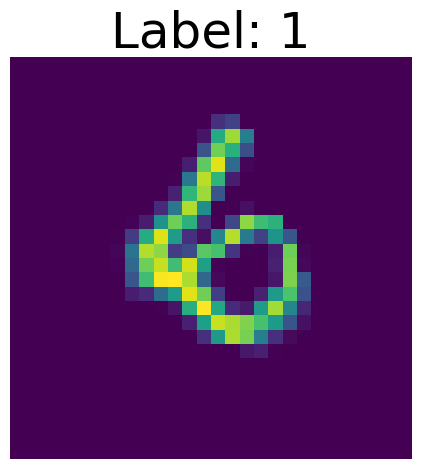} &
        \includegraphics[width=0.10\textwidth]{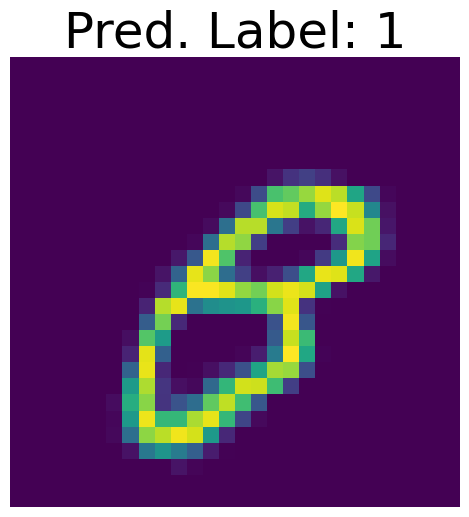} &
        \includegraphics[width=0.10\textwidth]{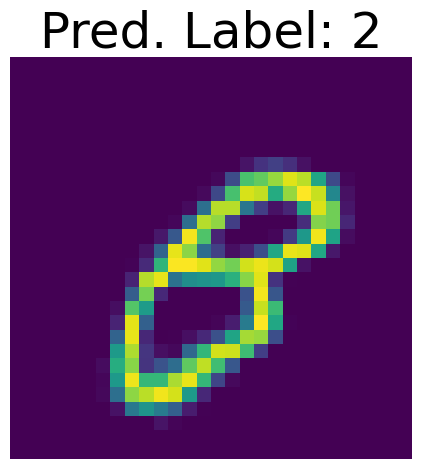} &
        \includegraphics[width=0.10\textwidth]{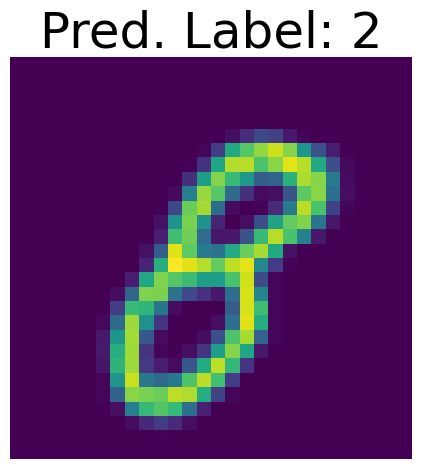} &
        \includegraphics[width=0.10\textwidth]{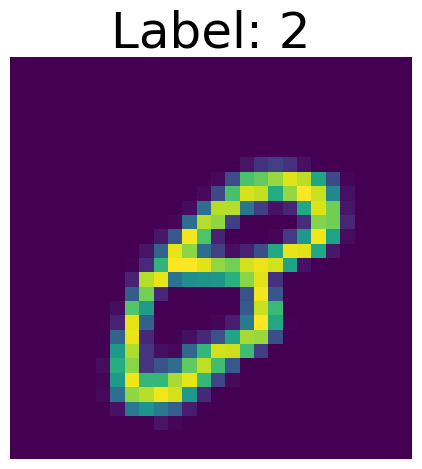} \\

        \includegraphics[width=0.10\textwidth]{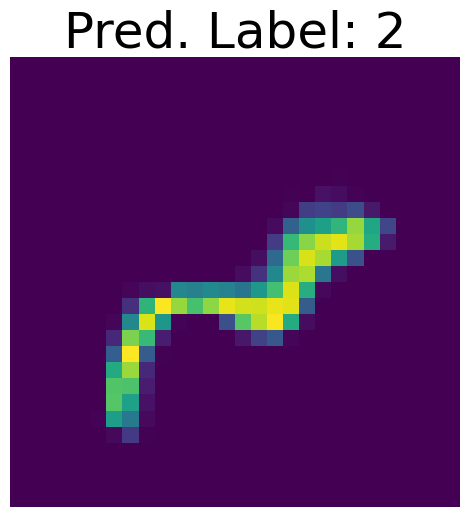} &
        \includegraphics[width=0.10\textwidth]{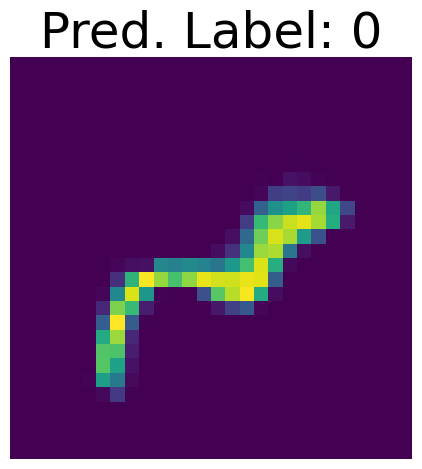} &
        \includegraphics[width=0.10\textwidth]{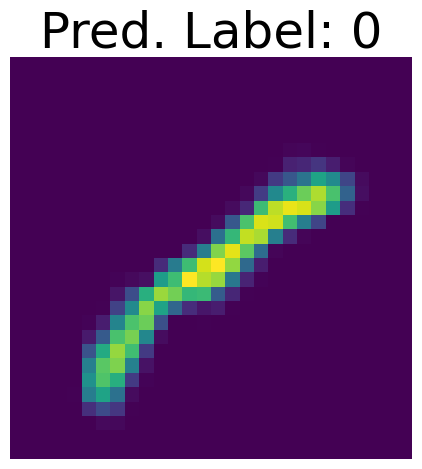} &
        \includegraphics[width=0.10\textwidth]{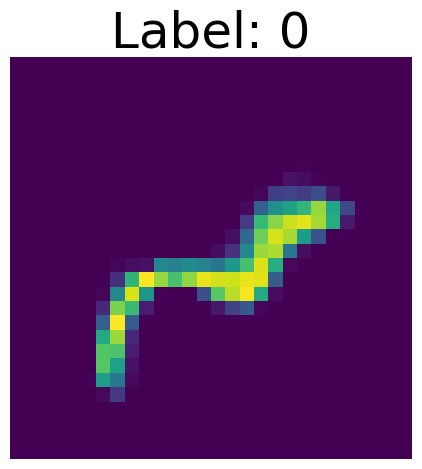} &
        \includegraphics[width=0.10\textwidth]{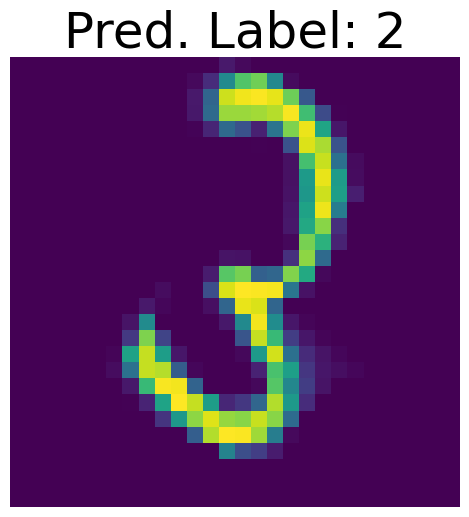} &
        \includegraphics[width=0.10\textwidth]{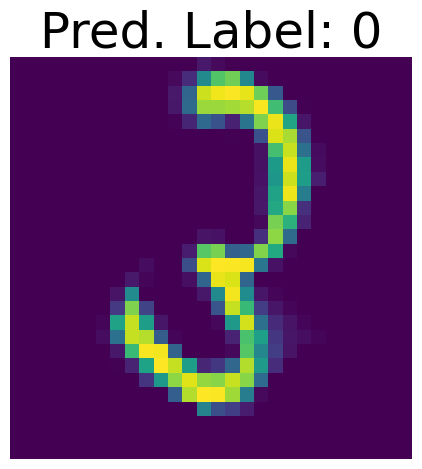} &
        \includegraphics[width=0.10\textwidth]{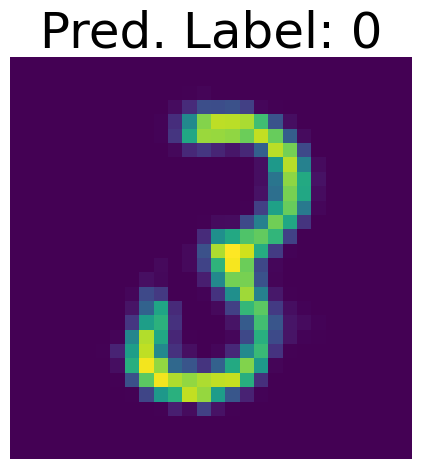} &
        \includegraphics[width=0.10\textwidth]{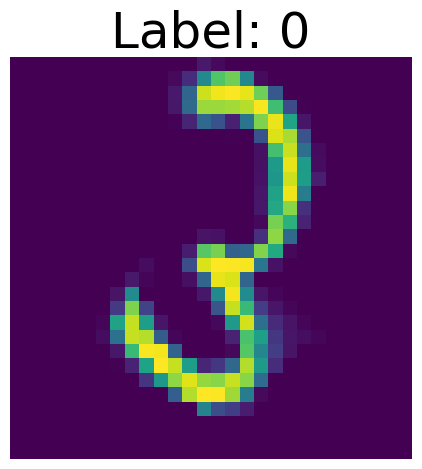} \\

        \includegraphics[width=0.10\textwidth]{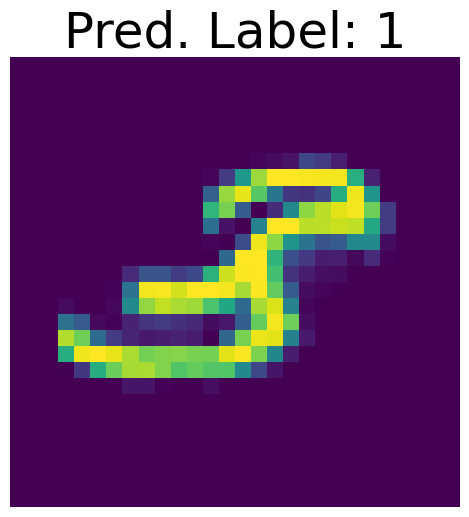} &
        \includegraphics[width=0.10\textwidth]{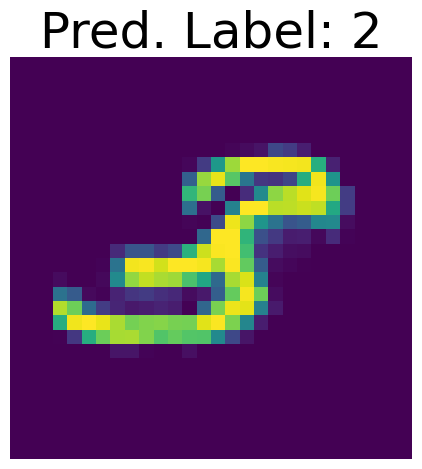} &
        \includegraphics[width=0.10\textwidth]{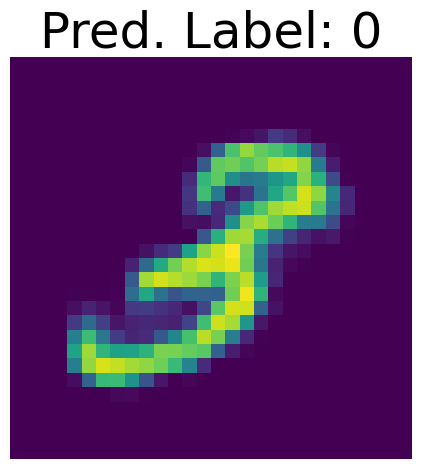} &
        \includegraphics[width=0.10\textwidth]{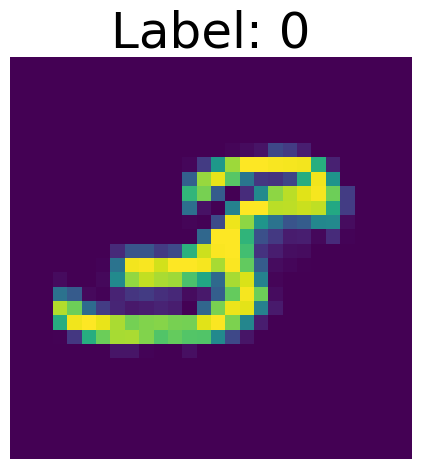} &
        \includegraphics[width=0.10\textwidth]{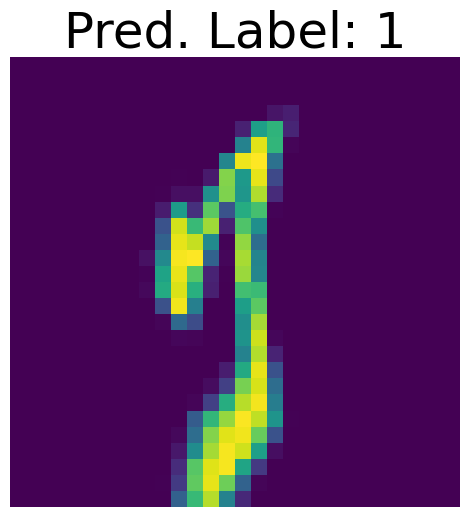} &
        \includegraphics[width=0.10\textwidth]{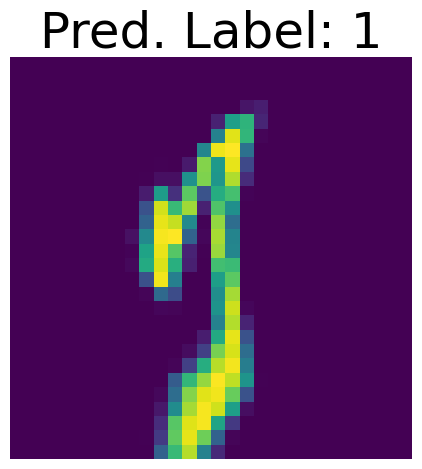} &
        \includegraphics[width=0.10\textwidth]{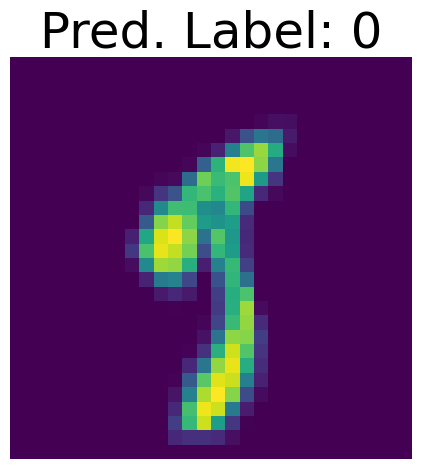} &
        \includegraphics[width=0.10\textwidth]{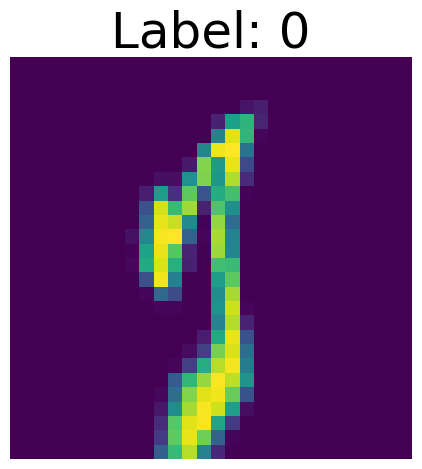}
         \\

        \includegraphics[width=0.10\textwidth]{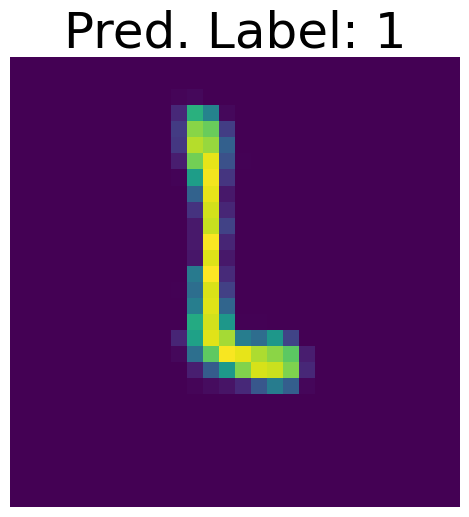} &
        \includegraphics[width=0.10\textwidth]{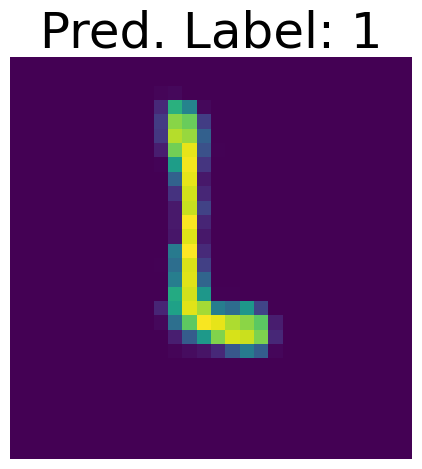} &
        \includegraphics[width=0.10\textwidth]{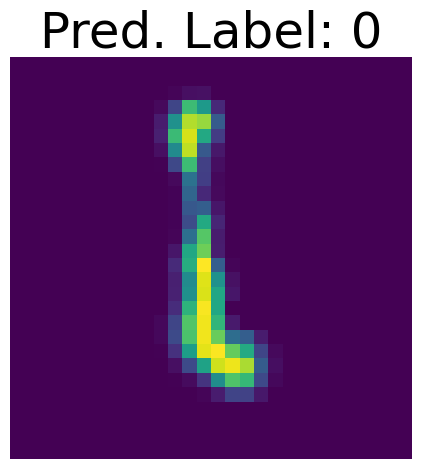} &
        \includegraphics[width=0.10\textwidth]{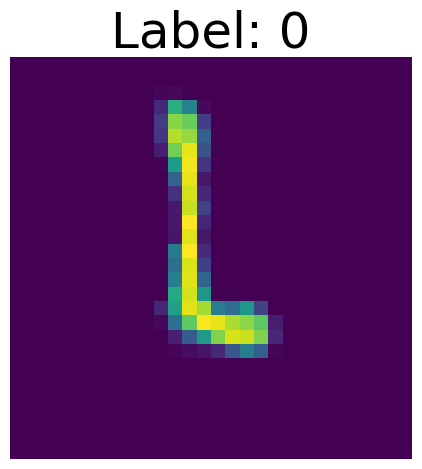} &
        \includegraphics[width=0.10\textwidth]{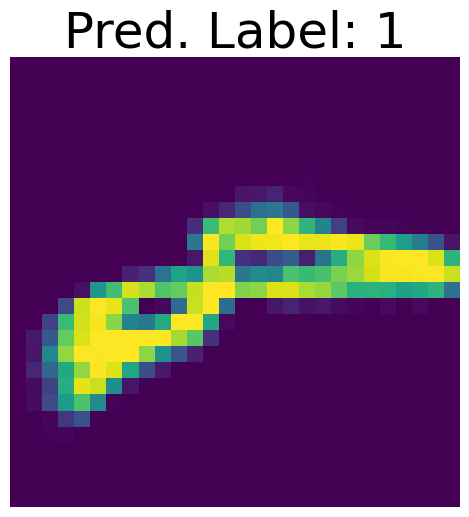} &
        \includegraphics[width=0.10\textwidth]{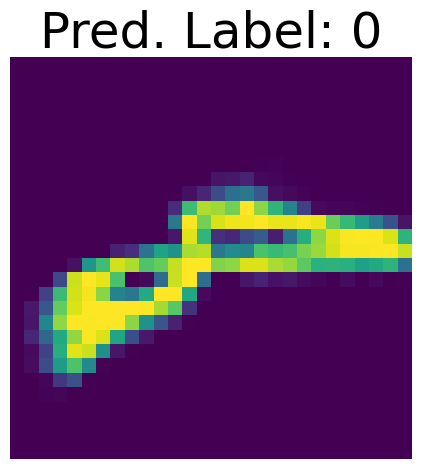} &
        \includegraphics[width=0.10\textwidth]{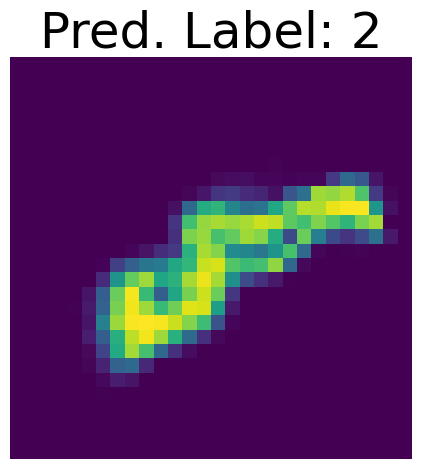} &
        \includegraphics[width=0.10\textwidth]{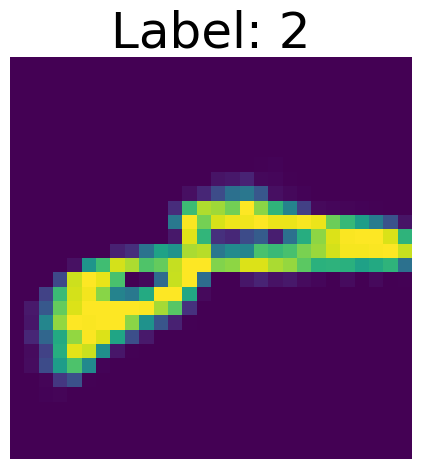}\\

        \includegraphics[width=0.10\textwidth]{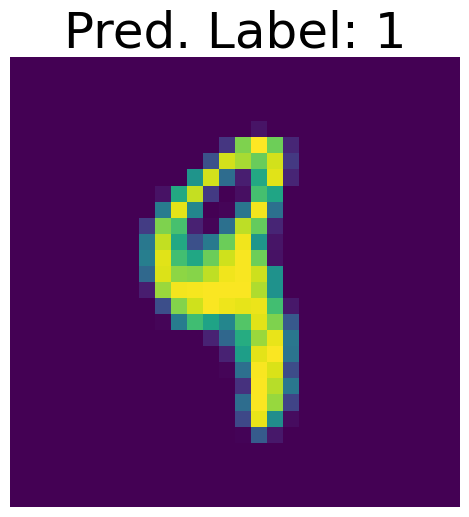} &
        \includegraphics[width=0.10\textwidth]{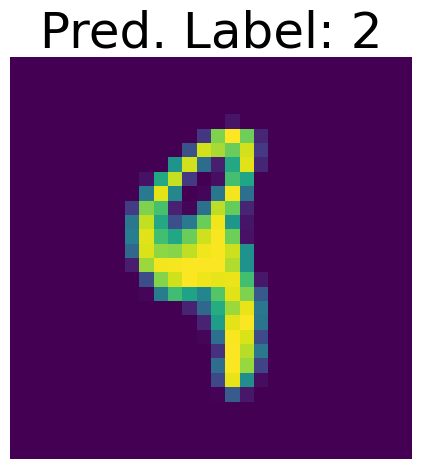} &
        \includegraphics[width=0.10\textwidth]{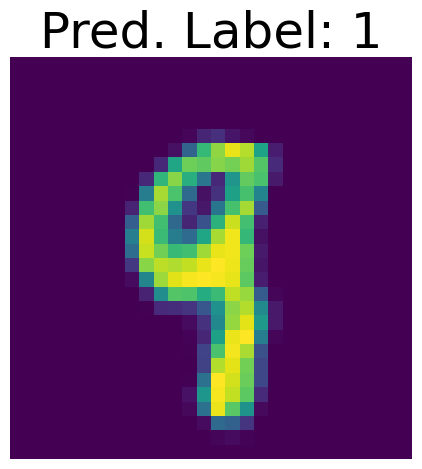} &
        \includegraphics[width=0.10\textwidth]{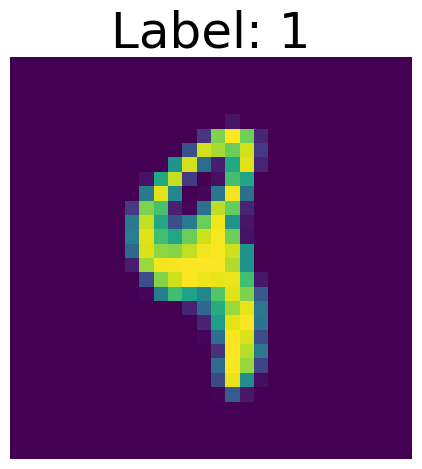} &
        \includegraphics[width=0.10\textwidth]{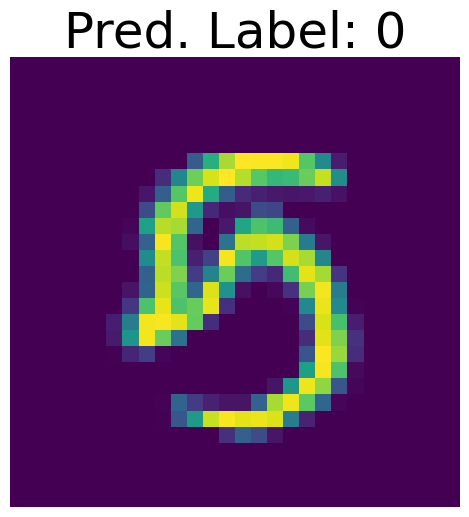} &
        \includegraphics[width=0.10\textwidth]{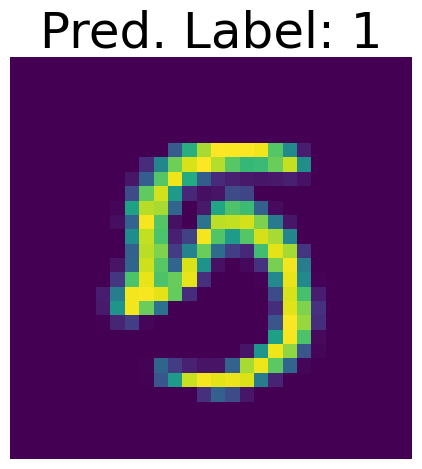} &
        \includegraphics[width=0.10\textwidth]{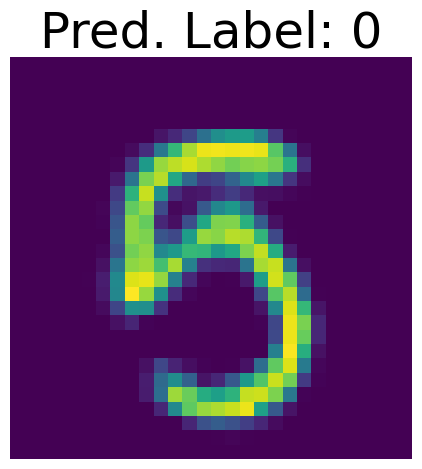} &
        \includegraphics[width=0.10\textwidth]{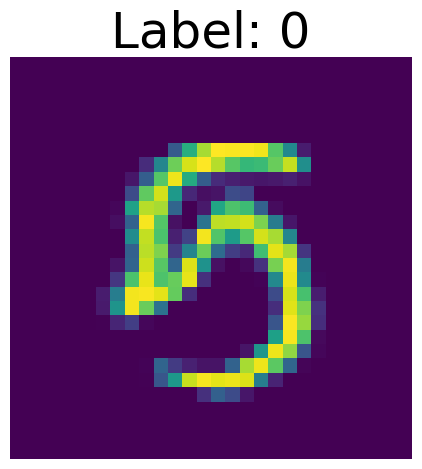} \\

    \end{tabular}
    \caption{Homology clas\-si\-fi\-cation task on $100$ diffeo\-morphi\-cally trans\-formed MNIST~\cite{mnist}.
    We show the input image with the predicted genus by the na\"ive approach \textbf{(column one)}, the augmented approach \textbf{(column two)}, and the ground truth \textbf{(column four)} as well as the canonicalised input and the prediction by DiffeoNN \textbf{(column three)}.
    For DiffeoNN and the na\"ive approach, we train a simple convolutional classifier for determining the genus, an adversarial network and an VAE with latent space dimension $10$ on the original MNIST.
    The augmented classifier is trained similarly but on a dataset containing original and transformed MNIST images.
    In DiffeoNN, we use $100$ gradient steps in the canonicalisation and set $\lambda_{\text{adv}}= 0.01$, $\lambda_{\text{VAE}} = 0.01$, $\lambda_{\Delta} = 1$, and $\lambda_{\mathcal{J}} = 10$.
    The canonicalised images in DiffeoNN are visually``closer'' to the original MNIST.
    The na\"ive classifier struggles most with the transformed input images, but also the augmented approach predicts wrong labels, see last three rows.
    }
\end{figure}

\end{document}